\documentclass{article} 
\usepackage{iclr2027_conference,times}

\usepackage{amsmath,amsfonts,bm}

\def\eqref#1{equation~\ref{#1}}

\def\1{\bm{1}}

\DeclareMathAlphabet{\mathsfit}{\encodingdefault}{\sfdefault}{m}{sl}
\SetMathAlphabet{\mathsfit}{bold}{\encodingdefault}{\sfdefault}{bx}{n}

\usepackage{url}
\usepackage{booktabs}
\usepackage{multirow}
\usepackage{graphicx}
\usepackage{array}
\usepackage{float}
\usepackage{placeins}
\usepackage{amssymb}
\usepackage{fvextra}
\usepackage{hyperref}

\title{InstanceBench: Diagnosing Referential Reasoning and Target Identity in Referring Expression Segmentation}

\author{
\textbf{Yuchen Li}$^{1}$ \quad
\textbf{Shaoyang Zhou}$^{1}$ \quad
\textbf{Yiran Wang}$^{1}$ \quad
\textbf{Ruiyi Deng}$^{1}$ \quad
\textbf{Haoyu Wang}$^{1}$ \\
\textbf{Ziru Wei}$^{2}$ \quad
\textbf{Zhen Zhao}$^{3}$ \quad
\textbf{Luping Zhou}$^{1}$ \\[0.5em]
{\small $^{1}$The University of Sydney \quad
$^{2}$The ATLAS Institute \quad
$^{3}$Shanghai Artificial Intelligence Laboratory} \\[0.4em]
{\small Project page: \url{https://github.com/aierwaixi/InstanceBench}}
}

\iclrpreprintcopy
\begin{document}

\maketitle

\begin{abstract}
Referring Expression Segmentation (RES) links natural-language descriptions to pixel-level object masks. Yet standard evaluation provides limited insight into instance-level referential reasoning: it does not systematically distinguish referential logics, test target preservation across valid grounding paths, or separate target-selection from mask-generation errors. We introduce InstanceBench, an instance-centered diagnostic benchmark comprising 6,194 images, 9,264 target instances, and 25,077 human-verified expressions. Each \emph{target-centric expression set} (TCES) fixes the image and target mask while pairing a minimal expression with a same-target variant that uses another valid cue or grounding path. A compact referential-logic taxonomy spans direct target evidence, same-class selection, relational and compositional grounding, and exclusion, while logic-critical construction suppresses simpler shortcuts. Identity-aware metrics measure target retention and set-level success while separating selection from mask-generation errors. Across 22 native-mask RES checkpoints from 18 model families, the strongest checkpoint reaches 67.1\% mIoU but only 59.6\% All@0.7. Controlled interventions confirm language sensitivity, while failure decomposition identifies target selection rather than mask decoding as the main bottleneck. On a controlled training subset, matched supervision improves identity-aware performance, showing that the diagnosed capability responds to targeted supervision. Collectively, InstanceBench supports a measure–diagnose–improve cycle: measuring target consistency across grounding paths, localizing failure sources, and evaluating targeted interventions.
\end{abstract}

\section{Introduction}

Referring Expression Segmentation (RES) produces a pixel-level mask for the object instance specified by a natural-language expression. Unlike category-level segmentation, RES requires two coupled decisions: target selection and mask generation. This distinction is critical when multiple objects of the same category share a scene. For example, the same person may be identified as ``the second person from the left,'' ``the person beside the bicycle,'' or ``the person between two seated women''---three valid expressions that follow different grounding paths but should return to the same target instance. Rather than mere paraphrases, these expressions instantiate different \emph{referential logics}: ways of structuring visual evidence to distinguish the target from competing instances. Executing these referential logics requires more than image–text alignment: models must identify target evidence, compare same-class instances, resolve object relations, compose multiple constraints, or eliminate alternatives.

Datasets such as RefCOCO, RefCOCO+, and RefCOCOg~\citep{refcoco,refcocog} have advanced RES, and recent systems~\citep{xsam,x2sam,hyperseg,instructseg} achieve increasingly strong mask-overlap scores. Yet it remains unclear how reliably these models execute different forms of instance-level referential reasoning. Standard evaluation provides limited diagnostic insight: expressions are not systematically organized by the logic needed to identify the target; alternative valid descriptions of the same instance are evaluated independently, obscuring consistency across grounding paths; and mask overlap conflates target-selection with mask-generation errors. 

Recent benchmarks (Table~\ref{tab:benchmark_comparison}) broaden referring-expression evaluation through controlled
and compositional expressions (CLEVR-Ref+~\citep{clevrref} and
Cops-Ref~\citep{copsref}), fine-grained compositional grounding with hard
negatives (FineCops-Ref~\citep{finecopsref,liu2024finecops}), generalized or
reasoning-intensive references (gRefCOCO~\citep{grefcoco}, MMR~\citep{mmr}, and
ReasonSeg~\citep{lai2024lisa}), ambiguity and shortcut diagnostics
(aRefCOCO~\citep{arefcoco} and Ref-Adv~\citep{refadv}), and counterfactual tests
of pixel-grounding hallucination (HalluSegBench~\citep{halluseg}). 
These benchmarks increase task difficulty and diagnostic coverage but primarily score expressions independently. Even when a target has several annotated expressions, as in RefCOCO, these are not systematically constructed to identify it through different referential logics, and their joint success is not a primary evaluation target. A complementary axis thus remains largely unmeasured: referential reliability, i.e., whether a model preserves a fixed target's identity across valid expressions that identify it through distinct, logic-critical grounding paths. Measuring it raises three instance-level questions: Which referential logics can models reliably execute? Do different valid grounding paths preserve the same target? When they fail, is the error in target selection or mask generation?
\begin{figure*}[t]
  \centering
  \includegraphics[width=\textwidth]{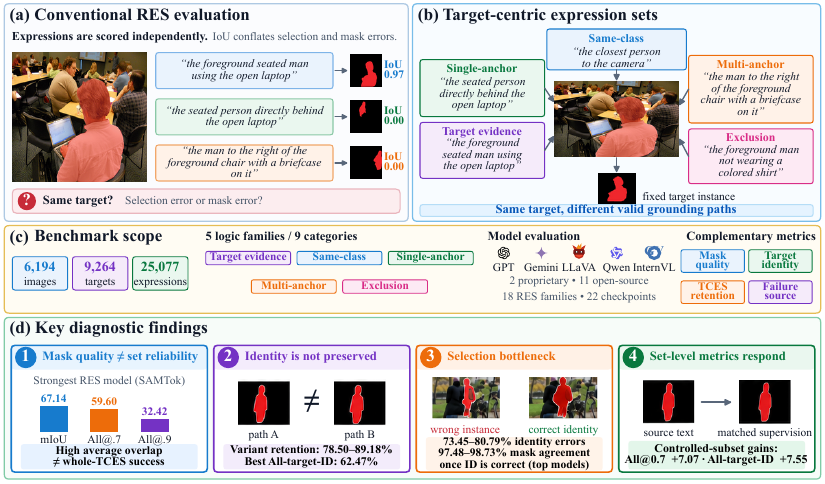}
  \vspace{-7mm}
  \caption{\textbf{Overview of InstanceBench.}
  (a) Independent RES scoring conflates target selection with mask quality.
  (b) A target-centric expression set (TCES) fixes the image and target while
  varying valid grounding paths. (c) Benchmark scope: the referential-logic taxonomy, evaluated models, and complementary metrics. (d) Diagnostics reveal reliability and identity gaps, localize failures to target selection, and assess matched supervision.}
  \label{fig:instancebench_overview}
  \vspace{-6mm}
\end{figure*}

To this end, we introduce \textbf{InstanceBench}, an instance-centered dataset and diagnostic benchmark comprising 6,194 COCO and OpenImages~\citep{kuznetsova2020open,benenson2019large} images, 9,264 target instances, and 25,077 human-verified expressions. Its basic unit is a \textbf{target-centric expression set (TCES)}: one image, one fixed target mask, and multiple valid expressions that refer to that same instance. Each core TCES pairs a minimal expression with an alternative that identifies the target through a different valid cue; additional annotations support controlled analyses of linguistic reformulation and target-preserving variation. InstanceBench categorizes expressions by the \emph{referential logic} required to isolate the target, rather than by surface wording. Its taxonomy spans direct target evidence, same-class set selection, single-anchor relations, multi-anchor or nested composition, and negation or exclusion. Paraphrases preserve both the underlying logic and grounding path, whereas cross-logic variants change the grounding path while keeping the image and target fixed. Logic-critical, shortcut-controlled construction ensures that the annotated logic is necessary for disambiguation. Identity-aware and TCES-level metrics measure which referential logics models can execute reliably, whether target identity is preserved across grounding paths, and whether failures arise from target selection or mask generation.
Across 22 native-mask RES checkpoints from 18 model families, the strongest one reaches 67.1\% expression-level mIoU but only 59.6\% All@0.7. Controlled text interventions establish language sensitivity, yet performance varies sharply across referential logics and models often fail to preserve the target across valid grounding paths. Failure decomposition identifies target selection as the dominant bottleneck in key relational and compositional failures. InstanceBench therefore provides an actionable map of current RES capability more than merely a harder test.

Our contributions are threefold: \textbf{(i)} we introduce \textbf{InstanceBench}, an instance-centered RES benchmark that systematically evaluates referential reasoning across nine logic categories. Its target-centric expression sets hold the image and target fixed while varying valid grounding paths, enabling controlled evaluation beyond isolated expression–mask pairs. \textbf{(ii)} we develop identity-aware and set-level diagnostics that measure target preservation across grounding paths and separate target-selection from mask-generation failures, complementing conventional mask-overlap metrics. \textbf{(iii)} through extensive evaluation of 22 native-mask RES checkpoints from 18 model families, we reveal substantial variation across referential logics and identify target selection as the dominant bottleneck in key relational and compositional failures. Controlled interventions and matched logic-aware supervision validate the diagnosis, improving identity-aware performance.

\begin{table*}[t]
  \centering
  \caption{Comparison with representative published REC/RES benchmarks. 
  }
  \label{tab:benchmark_comparison}
  \setlength{\tabcolsep}{2.4pt}
  \renewcommand{\arraystretch}{1.12}
  \scriptsize
  \resizebox{\textwidth}{!}{%
  \begin{tabular}{@{}l*{8}{c}@{}}
    \toprule
    & \multicolumn{2}{c}{Data}
        & \multicolumn{3}{c}{Diagnostic construction}
        & \multicolumn{3}{c}{Diagnostic evaluation} \\
    \cmidrule(lr){2-3}\cmidrule(lr){4-6}\cmidrule(lr){7-9}
    Benchmark
      & \shortstack{Real\\images}
      & \shortstack{Pixel\\masks}
      & \shortstack{Taxonomy\\scale}
      & \shortstack{Controlled\\construction}
      & \shortstack{Controlled same-target\\grounding-path variants}
      & \shortstack{Logic-wise\\diagnosis}
      & \shortstack{Set-level\\reliability}
      & \shortstack{Selection--mask\\decomposition} \\
    \midrule
    RefCOCO/+/g
      & $\surd$ & $\surd$ & -- & & & & & \\
    CLEVR-Ref+
      & & $\surd$ & 5 categories & $\surd$ & & $\surd$ & & \\
    Cops-Ref
      & $\surd$ & & 6 logic forms & $\surd$ & & $\surd$ & & \\
    FineCops-Ref
      & $\surd$ & & 6 structures & $\surd$ & $\surd$ & $\surd$ & & \\
    gRefCOCO
      & $\surd$ & $\surd$ & -- & & & & & \\
    ReasonSeg
      & $\surd$ & $\surd$ & -- & & & & & \\
    MMR
      & $\surd$ & $\surd$ & -- & & & & & \\
    Ref-Adv
      & $\surd$ & & -- & $\surd$ & & $\surd$ & & \\
    aRefCOCO
      & $\surd$ & $\surd$ & -- & $\surd$ & & $\surd$ & & \\
    HalluSegBench
      & $\surd$ & $\surd$ & -- & $\surd$ & & $\surd$ & & \\
    \midrule
    \textbf{InstanceBench (Ours)}
      & $\surd$ & $\surd$ & \textbf{9 categories} & $\surd$ & $\surd$
      & $\surd$ & $\surd$ & $\surd$ \\
    \bottomrule
  \end{tabular}%
  }
\end{table*}

\section{The InstanceBench Dataset}
\subsection{Data Sources and Splits}
InstanceBench contains 6,194 unique images, 9,264 target instances,
and 25,077 retained human-verified expressions. The
training, validation, and test sets contain 3,299/895/2,000 images, 4,896/1,450/2,918 TCESs, and 14,083/3,624/7,370 expressions, respectively. Images, category labels, and instance masks are drawn from the corresponding source splits of RefCOCO/+/g and OpenImages V7; all referring expressions are newly constructed and
human-verified. We apply the GE4 criterion, which requires at least four instances of the target category in each selected image. Because this criterion leaves the eligible RefCOCO/+/g pool strongly skewed toward person-category targets, we complement it with OpenImages, which contributes a broader range of object categories and same-class multi-instance scenes. We remove cross-source perceptual duplicates. The
resulting benchmark covers 221 normalized categories across the training,
validation, and test splits. Appendix Figure~\ref{fig:dataset_statistics}
summarizes the full-dataset target-category and referential-logic distributions.

\subsection{Dataset Design and Construction}

\noindent\textbf{Target-centric expression sets.}
Instead of evaluating expression--mask pairs independently, InstanceBench uses
the target instance and its coordinated expressions as the joint evaluation
unit. A target-centric expression set (TCES) is defined as
$\mathcal{T}=(I,G,\mathcal{E})$, where $I$ is an image, $G$ is a fixed ground-truth instance mask, and $\mathcal{E}=\{e_1,\ldots,e_K\}$ contains valid expressions that all refer to $G$. Each TCES contains two core expression roles. The \textbf{minimal}
expression contains only the cues needed to identify $G$ through its annotated
referential logic and grounding evidence, excluding descriptors that could
independently identify the target and bypass the intended grounding path. For
example, if one red-clothed woman appears among several men and color is the
intended identifying cue, we use ``the person in red'' rather than ``the woman
in red,'' because ``woman'' alone would already identify the target and
make the color cue unnecessary. The \textbf{same-target variant} expression refers to the
same target through a different valid cue or grounding path. Each expression
receives one canonical referential-logic label, and the two roles
preferentially use different referential logics whenever supported by the
image. A dense diagnostic subset additionally includes paraphrases that vary linguistic realization while preserving both referential logic and grounding path.

\begin{figure*}[t]
  \centering
  \includegraphics[width=0.95\textwidth]
  {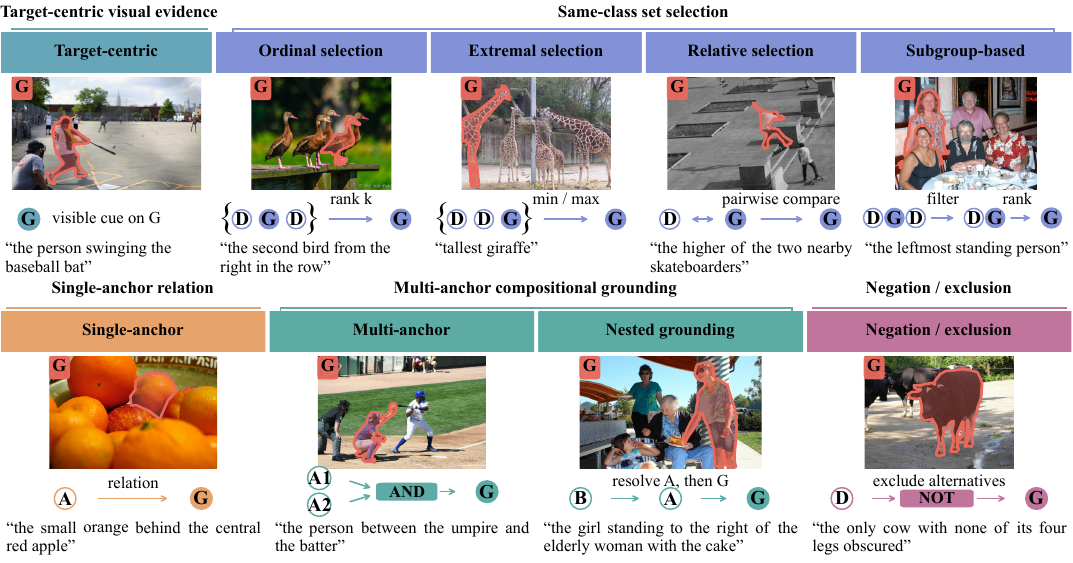}
  \vspace{-3mm}
  \caption{\textbf{Referential-logic taxonomy.}
  InstanceBench organizes nine categories into five evidence families.
  $G$, $D$, and $A/B$ denote the target, same-class distractors, and anchors;
  \textsc{and}, $B\!\rightarrow\!A\!\rightarrow\!G$, and \textsc{not} denote
  joint constraints, nested anchor resolution, and exclusion.}
  \label{fig:reasoning_taxonomy}
  \vspace{-5mm}
\end{figure*}

\noindent\textbf{Referential-logic taxonomy.}
We use \emph{referential logic} to denote how the visual evidence needed to
distinguish a target from competing instances is structured and resolved,
rather than the surface wording or cue types in an expression. A target may be
identified from evidence on the target itself, by selecting it within a
same-class set, through a relation to a single anchor, by jointly or recursively
resolving multiple anchors, or by excluding competing instances. These five
evidence structures induce nine canonical categories
(Figure~\ref{fig:reasoning_taxonomy}): \emph{target-centric visual evidence};
\emph{same-class set selection}, including ordinal, extremal, relative, and
subgroup-based selection; \emph{single-anchor relation}; \emph{multi-anchor
compositional grounding}, including parallel multi-anchor relation and
hierarchical nested grounding; and \emph{negation/exclusion}. A category is assigned only when its corresponding reasoning operation is
necessary for uniquely identifying the target: mentioning another object
is relational only when resolving that anchor is necessary, while negative
wording constitutes exclusion only when eliminating competing instances is
required. Appearance, action, possession, and fine-grained relation types are
retained as auxiliary metadata rather than separate logic categories. Complete
definitions, boundary cases, precedence rules, and additional examples are
provided in Appendix~\ref{app:taxonomy}.

\noindent\textbf{Logic-controlled generation.} 
For each GE4-qualified target, the image, target identity, and mask remain
fixed. Every expression must uniquely identify its target and be \emph{logic-critical} and \emph{shortcut-controlled}: its annotated
referential logic must be necessary, and no other descriptor may identify the
target independently. Given the image, target mask, and category, we use GPT-5.5~\citep{openai2026gpt55} in a two-stage prompting pipeline. Crucially,
GPT-5.5 neither defines the referential-logic taxonomy nor freely chooses an
expression's logic. We provide the taxonomy and validity contract; the analysis
stage identifies which prescribed logic--cue specifications are feasible
for the fixed target. A quota-aware scheduler selects two supported specifications while favoring logic diversity and under-represented logic categories; and the generation stage produces the minimal expression and same-target variant. Figure~\ref{fig:construction_two_pass} summarizes the pipeline, with complete prompts in Appendix~\ref{app:gpt_prompt_v22}.

\noindent\textbf{Human verification.} 
Every proposal undergoes two serial reviews by different annotators; seven
annotators participated across batches. Reviewer~1 checks mask quality, referential correctness, uniqueness,
naturalness, logic necessity, and taxonomy assignment, rejecting unusable masks
and directly revising correctable expressions. When a proposed expression could
plausibly refer to multiple instances, Reviewer~1 rewrites it to restore
uniqueness; Reviewer~2 then independently re-verifies the revised TCES for
uniqueness and logic necessity before finalization. Approved TCESs then undergo automated integrity checks. Of 9,885 resolved targets, 9,264 (93.7\%) were retained and 621 (6.3\%) rejected. Of 25,077 released expressions, 14,938 (59.6\%) derive from retained generated items and 10,139 (40.4\%) are human-added replacements or additions. 
A separate blinded audit evaluates taxonomy reproducibility and target validity
on 300 expressions from 150 image-unique TCESs. Two experienced annotators who
had not seen the sampled items independently select the referred same-class
candidate, assign one of the five
logic families, and assess uniqueness and logic necessity, without seeing the released target identities or taxonomy labels. They agree on
\textbf{91.7\%} of five-family assignments
(Cohen's $\kappa=\textbf{0.89}$; TCES-clustered 95\% CI: \textbf{88.3--94.7\%})
and on the selected target identity for
\textbf{95.7\%} of expressions. Their
target selections match the fixed target on \textbf{95.3\%} and
\textbf{98.3\%} of expressions, respectively. Appendix~\ref{app:independent_quality_audit} reports the complete protocol.
\begin{figure*}[t]
  \centering
  \includegraphics[width=\textwidth]
  {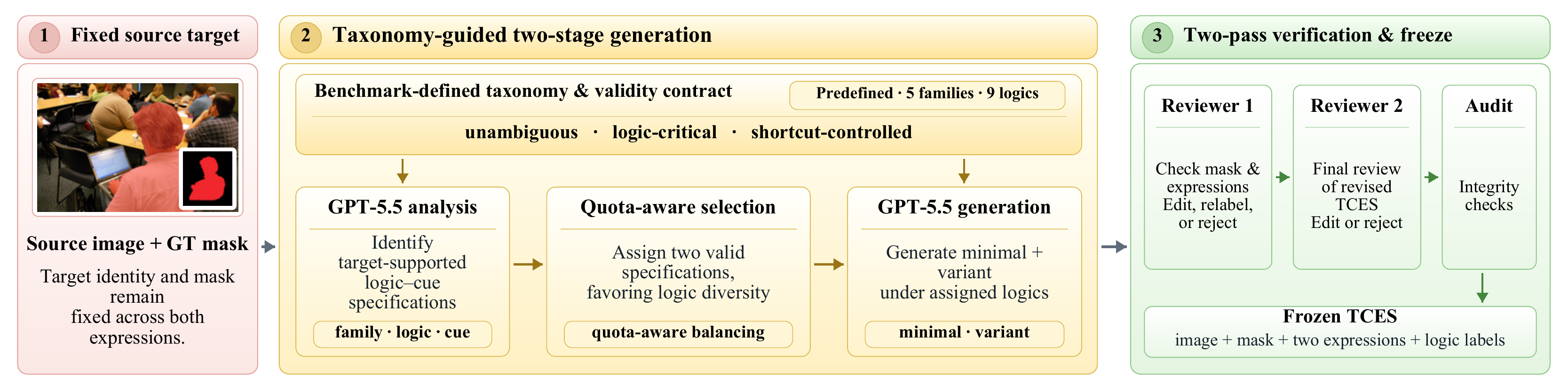}
  \vspace{-8mm}
  \caption{\textbf{InstanceBench construction pipeline.} 
}
  \label{fig:construction_two_pass}
  \vspace{-3mm}
\end{figure*}
\section{Evaluation Protocol}
InstanceBench evaluates expression-level mask quality, TCES-level reliability, and target-identity preservation. The headline metrics are
mIoU, All@0.7, and All-target-ID; the remaining measures diagnose retention, normalized consistency, and failure source. Let $\mathcal{T}$ denote the evaluated TCESs and $\mathcal{K}_t$ the expressions evaluated for TCES $t$. For expression $k$, $P_{t,k}$ is its predicted mask, $G_t$ is the fixed target mask, and $q_{t,k}=\operatorname{IoU}(P_{t,k},G_t)$. Unless otherwise stated, $\mathcal{K}_t$ contains the two core expressions; analyses over all available diagnostic expressions are explicitly marked as dense.

\noindent\textbf{Mask and set-level accuracy.}
At the expression level, we report the standard mIoU, oIoU, and Precision@$\tau$. At the TCES level, we define:
\begin{equation}
\mathrm{WorstPathIoU}=\frac{1}{|\mathcal{T}|}\sum_{t\in\mathcal{T}}
\min_{k\in\mathcal{K}_t}q_{t,k},\qquad
\mathrm{All@}\tau=\frac{1}{|\mathcal{T}|}\sum_{t\in\mathcal{T}}
\mathbf{1}\!\left[\min_{k\in\mathcal{K}_t}q_{t,k}\geq\tau\right].
\label{eq:set_level_metrics}
\end{equation}
Worst-path IoU exposes the weakest grounding path, whereas All@$\tau$ requires every evaluated expression in a TCES to succeed. We use $\tau\in\{0.7,0.9\}$.

\noindent\textbf{Target-identity assignment.}
For each TCES $t$, let
$\mathcal{C}_t=\{G_{t,1},\ldots,G_{t,n_t}\}$ denote all source-annotated
instances of the target category for which valid masks are available, including
the annotated target $G_t$. Identity-based evaluation uses
$\mathcal{T}_{\mathrm{ID}}=\{t\in\mathcal{T}:|\mathcal{C}_t|\geq2\}$.
To separate target selection from boundary quality, the evaluator assigns each
predicted mask $P$ to the most compatible $G_{t,i}\in\mathcal{C}_t$ using:
\begin{equation}
s(P,G_{t,i})=\frac{|P\cap G_{t,i}|}{\sqrt{|P||G_{t,i}|}}.
\label{eq:identity_assignment_score}
\end{equation}
Let $s_1$ and $s_2$ be the highest and second-highest scores. We accept the best match when $s_1\geq0.10$ and
$s_1-s_2\geq0.05$; the absolute threshold requires sufficient overlap, while
the margin rejects ambiguous assignments but tolerates boundary errors.
Otherwise, the identity is unresolved. Instances without source-provided masks
lie outside $\mathcal{C}_t$; a prediction matching no inventory instance
therefore remains unresolved and receives no identity credit. For each $P_{t,k}$,
we set $A_{t,k}=1$ only when the assigned instance is the annotated target of
TCES $t$; assignment to another instance or an unresolved result gives
$A_{t,k}=0$. Missing or invalid predictions receive zero IoU and unresolved
identity. Under the standard mask-output protocol, masks in $\mathcal{C}_t$ are used only by the evaluator and are never provided to the model.

\noindent\textbf{Correct identity preservation.}
On $\mathcal{T}_{\mathrm{ID}}$, we define \textbf{All-target-ID} as:
\begin{equation}
\mathrm{All\text{-}target\text{-}ID}
=\frac{1}{|\mathcal{T}_{\mathrm{ID}}|}
\sum_{t\in\mathcal{T}_{\mathrm{ID}}}
\mathbf{1}\!\left[A_{t,k}=1,\ \forall k\in\mathcal{K}_t\right].
\label{eq:all_target_id}
\end{equation}
Thus, agreement on the same wrong instance receives no credit. For a minimal expression $m$ and companion $c$, target-ID retention and target-and-mask retention are:
\begin{equation}
\mathrm{IDRet}=\Pr(A_{t,c}=1\mid A_{t,m}=1),\;
\mathrm{TM@}\tau=\Pr(A_{t,c}=1,q_{t,c}\geq\tau
\mid A_{t,m}=1,q_{t,m}\geq\tau).
\label{eq:retention_metrics}
\end{equation}
Target-ID retention measures robustness to an alternative valid expression, while TM@$\tau$ additionally requires both predictions to satisfy the mask threshold.

\noindent\textbf{Normalized consistency.}
Raw target-ID retention can be lower for a harder companion expression even
without weaker cross-expression coupling. We therefore compare a same-logic
paraphrase $p$ and an alternative-path variant $v$ on the same TCESs, sharing
the image, target, and minimal expression $m$. For $x\in\{p,v\}$, let
$\pi_m=\Pr(A_{t,m}=1)$, $\pi_x=\Pr(A_{t,x}=1)$, and
$\pi_{mx}=\Pr(A_{t,m}=1,A_{t,x}=1)$. For non-degenerate marginals, we define:
\begin{equation}
\mathrm{NCI}(m,x)=\frac{\pi_{mx}-\pi_m\pi_x}
{\min(\pi_m,\pi_x)-\pi_m\pi_x},\qquad
\Delta\mathrm{NCI}=\mathrm{NCI}(m,v)-\mathrm{NCI}(m,p).
\label{eq:normalized_consistency}
\end{equation}
NCI equals zero under marginal independence and one at the maximum co-success
permitted by the marginals; negative values indicate co-success below
independence. Hence, negative $\Delta\mathrm{NCI}$ indicates weaker
target-identity coupling after a grounding-path change than after same-logic
rewording.

\noindent\textbf{Failure localization.}
Predictions with $q_{t,k}<\tau$ are divided into \textbf{identity errors} ($A_{t,k}=0$) and \textbf{correct-ID/poor-mask errors} ($A_{t,k}=1$); confirmed switches to another same-category instance form a subset of identity errors. Finally, \textbf{conditional mask IoU}, $\operatorname{IoU}(P_{t,m},P_{t,c})$, is computed only when both predictions are assigned to the annotated target, isolating mask stability after correct target selection.

\section{Experiments}

\subsection{Experimental Setup}
\noindent\textbf{Evaluation split.}
All main results use the InstanceBench test set, comprising 2,000 images,
2,918 TCESs, and 5,836 core expressions, with one minimal expression and one
same-target variant per TCES. The test set also contains 1,534 additional
diagnostic expressions; analyses using these annotations are explicitly marked as dense.


\noindent\textbf{Models and inference.}
We evaluate 18 model families and 22 released checkpoint variants
(Table~\ref{tab:strict_overall}) using their official repositories, released
weights, and native image preprocessing. Universal multi-task systems (U) comprise SAMTok~\citep{zhou2026samtok}, X2SAM~\citep{x2sam}, UniPixel~\citep{liu2026unipixel}, Sa2VA~\citep{yuan2026sa2va}, X-SAM~\citep{xsam}, InstructSeg~\citep{instructseg}, HyperSeg~\citep{hyperseg}, and PSALM~\citep{zhang2024psalm}. RES models with additional reasoning or segmentation supervision (R+) comprise ConverSeg~\citep{sahoo2026conversational}, EVF-SAM2~\citep{zhang2026evf}, SaFiRe~\citep{arefcoco}, and SSP-SAM~\citep{tang2025ssp}; task-focused RES models (R) comprise DETRIS~\citep{huang2025densely}, PaDT~\citep{su2026patch}, VATEX~\citep{nguyen2025vision}, and LAVT~\citep{yang2022lavt}; CoHD~\citep{luo2025cohd} is the GRES specialist (G), and RefChess~\citep{tong2026refchess} is the training-free foundation-model composition (Z). Given their substantially broader training data, we use U models to estimate the current transfer-performance ceiling; causal architectural conclusions are based on the matched-supervision study in Section~\ref{sec:matched_training}. X2SAM and X-SAM main-leaderboard results average three full-test inference runs due to stochastic decoding; their dense identity, matched-logic, and intervention analyses use the pre-specified seed-0 run. All other checkpoints are deterministic in our setup.

\noindent\textbf{Statistical protocol.}
NCI is unitless; all other metrics are percentages. Pairwise comparisons and logic-category effects are clustered by TCES, so multiple expressions for the same target are not treated as independent samples. We report 95\% bootstrap confidence intervals for the central matched comparisons. All nine canonical referential-logic categories contain at least 113 expressions in the test set; smaller categories are interpreted with correspondingly wider uncertainty.

\subsection{Overall Performance: High mIoU Does Not Imply Set-Level Reliability}

\begin{table*}[t]
\centering
\caption{Overall performance on the InstanceBench core test set (2,918 TCESs;
5,836 expressions). All values are percentages; best results are bold.}
\label{tab:strict_overall}
\setlength{\tabcolsep}{3.6pt}
\renewcommand{\arraystretch}{1.03}
\scriptsize
\resizebox{0.97\textwidth}{!}{%
\begin{tabular}{llcrrrrrr}
\toprule
Model / checkpoint & Venue & Group & oIoU & mIoU & P@.7 & All@.7 & All@.9 & Worst-path IoU \\
\midrule
SAMTok-CO & CVPR'26 & U & \textbf{57.56} & \textbf{67.14} & \textbf{71.18} & \textbf{59.60} & 32.42 & \textbf{56.28} \\
X2SAM & ECCV'26 & U & 55.77 & 65.92 & 68.50 & 56.60 & \textbf{38.73} & 54.43 \\
UniPixel-7B & NeurIPS'25 & U & 54.68 & 62.78 & 63.18 & 51.82 & 24.61 & 52.07 \\
Sa2VA-2B & TPAMI'26 & U & 53.44 & 58.83 & 59.13 & 45.37 & 21.83 & 45.87 \\
X-SAM & AAAI'26 & U & 39.91 & 56.81 & 58.16 & 46.60 & 30.27 & 45.28 \\
InstructSeg & ICCV'25 & U & 41.55 & 52.73 & 53.51 & 39.07 & 21.38 & 38.52 \\
HyperSeg-3B & CVPR'25 & U & 41.49 & 52.58 & 53.80 & 39.75 & 22.72 & 38.93 \\
PSALM & ECCV'24 & U & 31.35 & 43.23 & 43.63 & 27.76 & 14.46 & 27.59 \\
ConverSeg-3B & CVPR'26 & R+ & 48.15 & 56.84 & 55.83 & 43.04 & 13.98 & 45.32 \\
EVF-SAM2 & IVC'26 & R+ & 46.75 & 55.37 & 54.76 & 39.07 & 18.57 & 40.53 \\
SaFiRe-mixed & NeurIPS'25 & R+ & 37.55 & 46.33 & 44.07 & 31.19 & 10.90 & 33.77 \\
SSP-SAM-RefCOCO & TCSVT'25 & R+ & 16.35 & 17.06 & 3.74 & 1.99 & 0.62 & 11.05 \\
SSP-SAM-RefCOCOg & TCSVT'25 & R+ & 16.59 & 16.78 & 3.50 & 1.99 & 0.69 & 10.86 \\
SSP-SAM-RefCOCO+ & TCSVT'25 & R+ & 15.87 & 16.34 & 3.34 & 1.82 & 0.58 & 10.57 \\
DETRIS & AAAI'25 & R & 32.89 & 44.54 & 44.29 & 30.09 & 10.76 & 30.95 \\
PaDT-REC-3B & ICLR'26 & R & 36.51 & 42.01 & 34.73 & 29.95 & 3.36 & 34.89 \\
VATEX-Swin-B & WACV'25 & R & 25.53 & 37.34 & 36.87 & 21.80 & 10.66 & 22.54 \\
LAVT-RefCOCO & CVPR'22 & R & 26.13 & 34.06 & 29.23 & 14.67 & 5.14 & 19.23 \\
LAVT-RefCOCOg & CVPR'22 & R & 28.56 & 33.31 & 25.51 & 12.95 & 3.77 & 19.71 \\
LAVT-RefCOCO+ & CVPR'22 & R & 25.17 & 32.40 & 28.72 & 15.08 & 5.14 & 18.72 \\
CoHD & ICCV'25 & G & 28.26 & 35.82 & 33.04 & 18.92 & 7.57 & 21.93 \\
RefChess-hybrid & ICML'26 & Z & 19.36 & 30.06 & 27.91 & 15.32 & 8.19 & 16.75 \\
\bottomrule
\end{tabular}}
\vspace{-5mm}
\end{table*}

Table~\ref{tab:strict_overall} reveals a substantial gap between expression-level mask quality and target-level reliability. SAMTok obtains the strongest mIoU at 67.14\%, yet solves both core expressions in only 59.60\% of TCESs at IoU 0.7; at IoU 0.9 this falls to 32.42\%. X2SAM has a similar profile (65.92 mIoU and 56.60 All@.7), while its higher All@.9 illustrates that model rankings can depend on whether average overlap, whole-set reliability, or high-precision masks are required. Together with the lower WorstPathIoU values, these results show that expression-level averages can obscure weak grounding paths within a TCES and motivate reporting expression- and TCES-level metrics together.

\noindent\textbf{General-purpose MLLM extension.}
We additionally evaluate general-purpose MLLMs under two diagnostic protocols.
In direct localization, each model predicts a box that is converted into a mask
by the same frozen SAM~3 backend~\citep{carion2026sam}; oracle candidate-ID
evaluation instead supplies the source-annotated candidate inventory
$\mathcal{C}_t$ and
isolates target selection from mask generation. Qwen3.8-27B reaches 69.83\%
mIoU and 63.48\% All@.7 in the box-to-mask setting, while Gemini-3.8-Flash
achieves 76.29\% expression-level target accuracy but only 65.17\%
All-target-ID under oracle candidate selection. Thus, strong MLLMs also exhibit
a substantial gap between individual grounding accuracy and same-target
reliability. Because these protocols use either a separate segmentation backend
or oracle candidate masks, they are diagnostic extensions rather than directly
comparable native-mask RES evaluations. Complete results are reported in
Appendix Tables~\ref{tab:mllm_direct} and~\ref{tab:mllm_candidate_id}.

\subsection{Which Referential Logics Challenge Current Models?}

\begin{figure*}[t]
  \centering
  \includegraphics[width=\textwidth]
  {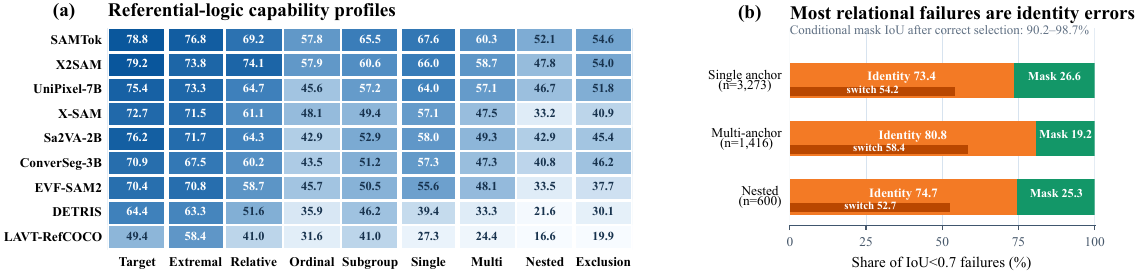}
  \vspace{-7mm}
  \caption{\textbf{Referential-logic capability and failure localization.}
  (a) Cells report expression-level mIoU across the nine canonical referential
  logics, revealing model-specific capability profiles. (b) At IoU 0.7,
  target-identity errors dominate failures on single-anchor, multi-anchor, and
  nested expressions. The dark inset denotes confirmed same-class switches, a
  strict subset of identity errors; the remaining green segment is correct-ID,
  poor-mask failure.}
  \label{fig:capability_failure}
  \vspace{-5mm}
\end{figure*}

Figure~\ref{fig:capability_failure}(a) reveals markedly non-uniform capability
profiles across referential logics. Direct target-centric evidence and extremal
descriptions are comparatively easy, while ordinal selection, negation/exclusion, and relational composition remain weaker. This variation is visible within every model and therefore cannot be summarized by a single scalar difficulty score. Complete numerical results and category counts are reported in Appendix Table~\ref{tab:logic_profile}. Per-model logic rankings are stable across COCO and OpenImages (per-model Spearman $\rho=0.75$--$0.95$; Appendix Figure~\ref{fig:source_logic_profiles} and Table~\ref{tab:source_logic_rank}). To reduce visual and linguistic confounding, we match ordinal and extremal examples by target size, category, expression length, and same-class instance count. Ordinal mIoU remains lower by 10.27 points for X2SAM (95\% CI $[-15.82,-4.89]$), 17.02 for SAMTok ($[-22.65,-11.53]$), 25.52 for UniPixel ($[-30.51,-20.29]$), and 20.86 for X-SAM ($[-26.23,-15.51]$). The ordinal--extremal gap therefore persists under these matched controls. Nested grounding also remains below matched spatial controls for X2SAM, SAMTok, and X-SAM, whereas the smaller multi-anchor--spatial effects are not statistically resolved for these models. Complete matched results appear in Appendix Table~\ref{tab:matched_logic_controls}. These results localize capability gaps to specific forms of referential binding.

\subsection{Same-Target Consistency and Failure Localization}
\begin{table*}[t]
\centering
\caption{\textbf{Same-target consistency.}
Retention/NCI use 988 COCO-derived matched TCESs; dense All-target-ID and core-pair
conditional mask IoU use 2,867 identity-eligible TCESs.
$\Delta$NCI is variant minus paraphrase; brackets are 95\% TCES-bootstrap CIs.}
\label{tab:identity_consistency}
\setlength{\tabcolsep}{3.0pt}
\renewcommand{\arraystretch}{1.03}
\scriptsize
\resizebox{\textwidth}{!}{%
\begin{tabular}{lccccccc}
\toprule
Model & Para. Ret. & Variant Ret. & Para. NCI & Variant NCI & $\Delta$NCI [95\% CI] & All-target-ID & Cond. mask IoU \\
\midrule
SAMTok & 97.55 & 88.47 & 0.565 & 0.151 & $-0.414$ [$-0.621,-0.242$] & 62.47 & 97.48 \\
X2SAM & 96.75 & 89.18 & 0.662 & 0.201 & $-0.462$ [$-0.634,-0.295$] & 58.32 & 98.05 \\
UniPixel & 97.42 & 89.16 & 0.770 & 0.287 & $-0.483$ [$-0.642,-0.322$] & 58.98 & 92.31 \\
Sa2VA & 96.17 & 85.43 & 0.593 & 0.147 & $-0.446$ [$-0.612,-0.288$] & 51.48 & 90.16 \\
ConverSeg & 96.11 & 87.22 & 0.660 & 0.340 & $-0.320$ [$-0.476,-0.169$] & 51.62 & 90.30 \\
PaDT & 96.07 & 83.84 & 0.644 & 0.185 & $-0.459$ [$-0.602,-0.335$] & 50.12 & 94.56 \\
X-SAM & 95.48 & 83.37 & 0.611 & 0.165 & $-0.446$ [$-0.583,-0.331$] & 48.45 & 98.73 \\
EVF-SAM2 & 94.31 & 78.50 & 0.583 & 0.058 & $-0.525$ [$-0.658,-0.395$] & 44.65 & 88.32 \\
\bottomrule
\end{tabular}}
\vspace{-4mm}
\end{table*}

Table~\ref{tab:identity_consistency} shows that paraphrase retention is
94.31--97.55\%, whereas alternative-path retention falls to
78.50--89.18\%, a drop of 7.58--15.81 points. Lower standalone variant accuracy does not explain this gap: paraphrase NCI is 0.565--0.770, compared with 0.058--0.340 for variants, and $\Delta$NCI ranges from $-0.320$ to $-0.525$, with every 95\% confidence
interval below zero. Thus, after matching the image, target, and minimal
expression and normalizing marginal accuracy, changing the grounding evidence
is associated with substantially weaker target-identity preservation than
same-logic rewording. Even SAMTok selects the annotated target across every dense expression in only 62.47\% of identity-eligible TCESs.
All-target-ID directly measures correct identity preservation across the set;
All@.7 additionally requires a good mask. Full track-specific retention and
TM@.7 results are reported in Appendix Table~\ref{tab:identity_tracks_full},
with a descriptive visualization in Appendix
Figure~\ref{fig:supplementary_retention_learnability}(a).

Figure~\ref{fig:capability_failure}(b) and Appendix
Table~\ref{tab:failure_decomp} decompose predictions below IoU 0.7 on
single-anchor and compositional expressions. Across five universal
models, target-identity errors account for 73.45--80.79\% of these failures;
confirmed same-class switches, a strict subset, already comprise
52.67--58.40\% of these failures. In a separate conditional analysis, when both
predictions are assigned to the annotated target, their mask IoU remains
90.16--98.73\%. Together, these results identify the primary bottleneck as
resolving and preserving target identity under relational constraints, rather
than mask delineation once the target is correctly resolved.
\begin{figure}
  \centering
  \includegraphics[width=0.95\textwidth]
  {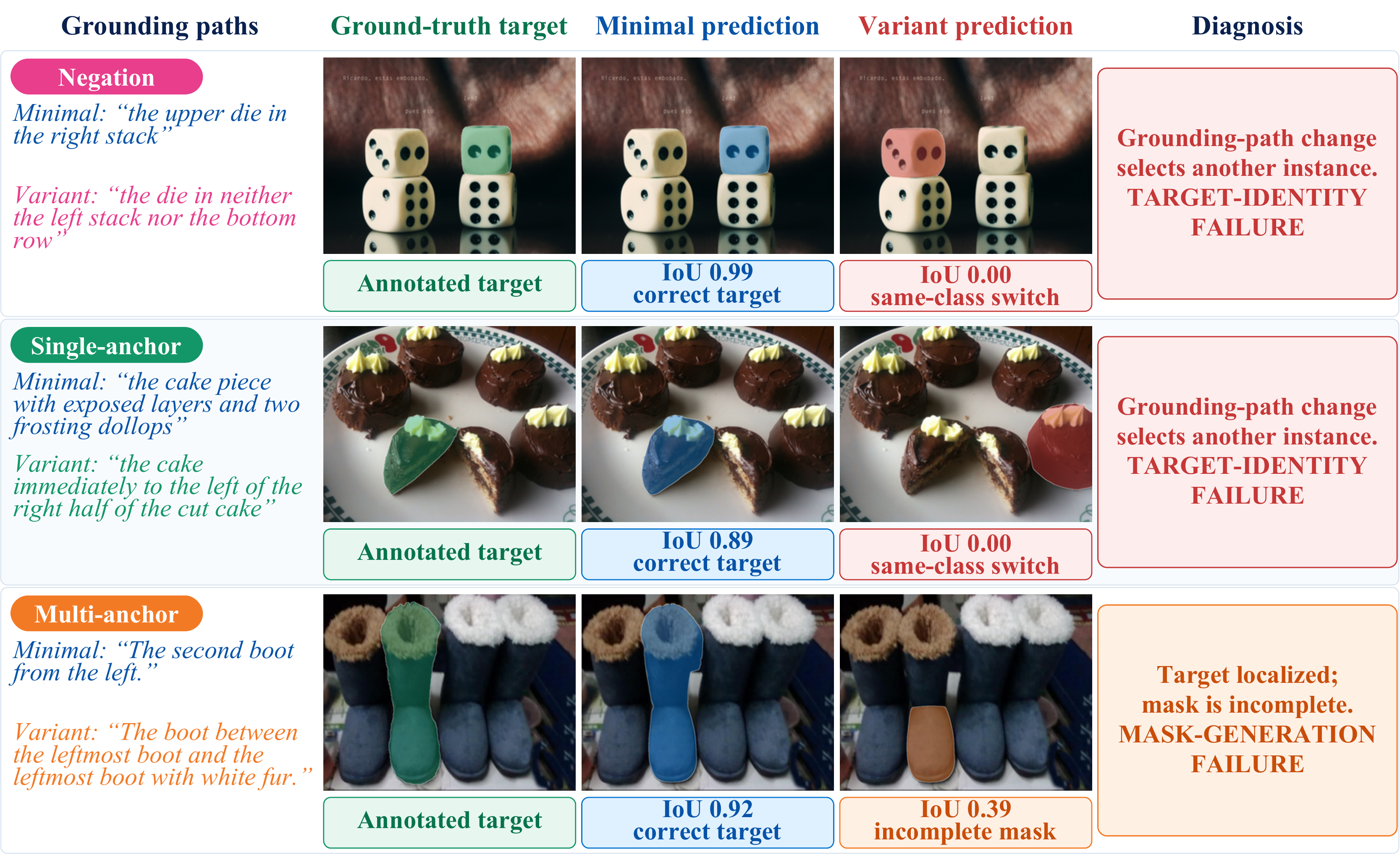}
  \vspace{-3mm}
  \caption{\textbf{Qualitative diagnoses under a fixed target.}
  Green: ground truth; blue: minimal prediction; red: wrong-instance variant; orange: incomplete target mask.
  }
  \label{fig:qualitative_diagnosis}
  \vspace{-5mm}
\end{figure}
\subsection{Language Sensitivity Is Not Referential Reliability}

We use two controlled interventions to test whether the failures arise from text neglect. Replacing each expression with its category name reduces mIoU by 36.95--44.91 points, showing that models rely on disambiguating language. More decisively, when an expression is replaced by one referring to another same-category instance in the same image, 90.29--94.23\% of predictions shift toward the newly described target. Systems are therefore language-responsive, but this sensitivity does not imply reliable execution of the required referential logic or preservation of target identity across valid grounding paths. Appendix Table~\ref{tab:interventions_full} reports full statistics.

\subsection{The Diagnosed Capability Responds to Supervision}
\label{sec:matched_training}

Appendix Figure~\ref{fig:supplementary_retention_learnability}(b) and
Table~\ref{tab:matched_training} evaluate whether the diagnosed capability
responds to targeted supervision. To isolate expression supervision from
source-domain shift, we use a fixed COCO-derived subset (930 training images;
1,980 TCESs) aligned with the RefCOCO-family domain of both architectures.
Images, masks, rows, initialization, optimization, and seeds are matched across
supervision settings. Relative to the source-text
control, logic-aware supervision improves LAVT mIoU, All@.7, and All-target-ID
by $5.00$, $7.07$, and $7.55$ points, respectively. DETRIS benefits most from
the 50/50 mixture, improving the same metrics by $1.78$, $2.55$, and $4.34$
points. The larger
target-level gains show that the diagnosed deficiency is learnable, while the
different best supervision setting reveals an architecture--supervision
interaction.
\subsection{Qualitative Diagnosis}

Figure~\ref{fig:qualitative_diagnosis} presents representative cases under a
fixed annotated target. Here, Minimal and Variant are two valid expression
roles for that target, not different model variants. In the first two rows,
changing the grounding path redirects the prediction to another same-category
instance despite an accurate minimal-expression prediction. In the third, the
model retains the intended instance but produces an incomplete mask. These
examples illustrate the two diagnosed mechanisms rather than their prevalence,
which is quantified in Figure~\ref{fig:capability_failure}(b) and Appendix
Table~\ref{tab:failure_decomp}.

\section{Conclusion}
InstanceBench reframes RES evaluation around target-identity reliability across valid grounding paths. On the test set, average mIoU overstates whole-TCES reliability, and errors arise mainly from target selection rather than mask generation. Counterfactual text interventions show that models are language-sensitive yet fail to preserve identity across valid grounding paths, while the controlled-subset study shows that this capability responds to supervision. As a controlled diagnostic benchmark rather than an estimate of RES performance under natural distributions, InstanceBench maps model capability over referential logics to guide targeted improvement.

\bibliography{iclr2027_conference}
\bibliographystyle{iclr2027_conference}

\clearpage
\appendix
\section{Expanded Related Work}
\label{app:related_work}

Table~\ref{tab:benchmark_comparison} summarizes the design axes that distinguish
InstanceBench from representative REC and RES benchmarks; here we discuss the
corresponding research lines in greater detail.

\noindent\textbf{Standard referring-expression benchmarks.}
RefCOCO, RefCOCO+, and RefCOCOg established the principal real-image test beds
for referring-expression comprehension and segmentation~\citep{refcoco,refcocog}.
RefCOCO and RefCOCO+ were collected through an interactive reference game,
with RefCOCO+ disallowing absolute-location terms, while RefCOCOg contains
longer descriptions collected in a non-interactive setting.  These datasets
provide diverse human expressions and instance annotations, but conventionally
evaluate each expression--target pair independently; they neither organize
expressions by required referential logic nor treat coordinated descriptions of
one fixed target as a set-level reliability unit.

\noindent\textbf{Controlled and compositional reference.}
CLEVR-Ref+ uses synthetic scenes, functional programs, and exact masks to expose
intermediate compositional reasoning steps under controlled visual
distributions~\citep{clevrref}.  Cops-Ref transfers controlled composition to
real images through an expression engine spanning six reasoning forms and a
test protocol with visually similar distracting images~\citep{copsref}.
FineCops-Ref further stresses fine-grained composition with controllable
difficulty, multi-hop relations, and edited negative text or image cases that
test whether a model can reject an absent referent~\citep{liu2024finecops,finecopsref}.
These benchmarks diagnose compositional grounding, whereas InstanceBench asks a
different question: whether distinct, valid, logic-critical grounding paths in
the same image return to the same annotated target.

\noindent\textbf{Generalized and reasoning segmentation.}
gRefCOCO extends conventional single-target RES to single-target, multi-target,
and no-target expressions with pixel supervision~\citep{grefcoco}.
ReasonSeg instead uses implicit instructions requiring semantic or world
knowledge to identify a mask~\citep{lai2024lisa}, while MMR expands reasoning
segmentation to multi-target outputs and both object- and part-level
granularity~\citep{mmr}.  These resources broaden what an expression may denote
and what knowledge it may require.  InstanceBench remains focused on
instance-level reference, but controls the image and target identity so that
changes across expressions can be attributed to the grounding path rather than
to a change in the requested output.

\noindent\textbf{Ambiguity, shortcuts, and counterfactual grounding.}
Ref-Adv suppresses shortcut cues using minimally sufficient expressions and
hard same-image distractors, and evaluates reasoning facets such as
negation~\citep{refadv}.  aRefCOCO reannotates challenging cases to probe
object-distracting and category-implicit ambiguity in referring image
segmentation~\citep{arefcoco}.  HalluSegBench pairs factual scenes with
controlled visual counterfactuals, testing whether segmentation models abstain
when the queried evidence has been removed or replaced~\citep{halluseg}.
InstanceBench is complementary to these stress tests: it holds the visual input
and ground-truth mask fixed, varies the valid evidence used to identify that
mask, and evaluates logic-wise capability, same-target set reliability, and the
separation of target-selection from mask-generation failures.
\begin{table}[t]
\centering
\caption{The complete InstanceBench referential-logic taxonomy. Categories are
defined by the evidence and inference required to uniquely identify the target.
An en dash indicates that a family is not further divided into subtypes.}
\label{tab:complete_taxonomy}
\setlength{\tabcolsep}{4pt}
\renewcommand{\arraystretch}{1.02}
\small
\resizebox{\textwidth}{!}{%
\begin{tabular}{p{3.0cm}p{2.5cm}p{5.1cm}p{5.0cm}}
\toprule
Family & Subtype & Example & Identification requirement \\
\midrule

Same-class set selection
& Ordinal
& ``the second wine bottle from the left''
& Order the same-class instances and select a specified index. \\

& Extremal
& ``the leftmost visible person''
& Select the instance attaining a positional or visual extremum. \\

& Relative selection
& ``the larger of the two pizza slices''
& Compare same-class instances using a visible property or distance. \\

& Subgroup-based
& ``the right pale-pink cruller among the two at the bottom''
& Form a constrained subgroup before selecting one member from it. \\

\midrule

Single-anchor relation
& --
& ``the young girl immediately to the left of the woman cutting the pizza''
& Resolve one independently identifiable anchor and apply one spatial, distance, contact, containment, facing, looking, or body-orientation relation. \\

\midrule

Multi-anchor compositional grounding
& Multi-anchor relation
& ``the dark suitcase between the wooden crate and the tan suitcase''
& Jointly resolve at least two anchors whose combined constraints identify the target. \\

& Nested grounding
& ``the woman to the left of the light-clothed man holding a wine glass''
& Resolve a relationally described anchor before using it to locate the target. \\

\midrule

Target-centric visual evidence
& --
& ``the foreground woman cutting the pizza with a fork and knife''
& Identify the target using its visible appearance, part, state, pose, action, possession, or interaction. \\

\midrule

Negation or exclusion
& --
& ``the person who is not wearing a T-shirt or a sleeveless shirt''
& Eliminate candidate instances satisfying the excluded condition. \\

\bottomrule
\end{tabular}}
\end{table}
\section{Taxonomy Annotation Guide}
\label{app:taxonomy}

\paragraph{Same-class set selection.}
This family explicitly constructs a set of same-category candidates and selects
one using ordinal, extremal, relative, or subgroup-based logic. For example,
``the second wine bottle from the left'' requires ordering the bottles before
indexing the target. By contrast, ``the person on the left side of the image''
is a scene-frame spatial description unless it denotes the leftmost member of a
same-class set.

\paragraph{Single-anchor relation.}
The target is identified through one relation to one independently resolvable
anchor. Spatial, distance, contact, containment, facing, looking, and
body-orientation relations share this evidence topology and are retained as
auxiliary relation types. Merely mentioning another object is insufficient:
resolving the anchor and applying the target--anchor relation must be necessary
to identify the target. Intrinsic pose or orientation that identifies the target
without resolving an anchor is instead target-centric visual evidence.

\paragraph{Multi-anchor compositional grounding.}
This family contains two distinct reasoning structures. A multi-anchor relation
jointly applies parallel constraints from multiple independently resolvable
anchors. In ``the dark suitcase between the wooden crate and the tan suitcase,''
both anchors constrain the target. Nested grounding instead follows a
hierarchical chain: it first resolves a relationally described anchor and then
uses that anchor to locate the target. If one independently resolvable anchor
alone suffices, the expression remains in the single-anchor family.

\paragraph{Target-centric visual evidence.}
The target is selected using its directly visible appearance, parts, state,
pose, action, possession, or interaction. This family provides a direct-evidence
reference condition against which set-based and relational logics can be
compared. A visible interaction remains target-centric when resolving the
interacted object as an independent anchor is unnecessary for identification.

\paragraph{Negation or exclusion.}
The target is identified by eliminating candidates for which an attribute,
object, action, or relation holds. Negation is assigned only when the excluded
condition is necessary for disambiguation; a redundant negative modifier does
not make an expression an exclusion example. Because elimination may operate on
target cues, relations, or candidate sets, exclusion is treated as a
cross-cutting logic but as one canonical evaluation category.

\section{Independent Quality-Audit Protocol}
\label{app:independent_quality_audit}

The independent audit uses a quota-balanced stratified sample of 300 core
expressions from 150 image-unique TCESs. The sample contains 75
COCO/RefCOCO-family and 75 OpenImages images and is balanced across the
benchmark's canonical referential logics. Two experienced annotators
independently review every expression; neither has previously seen any selected
image or expression.
For each item, the interface shows the original image and all annotated
same-class candidates under target-independent randomized identifiers while
withholding the released target identity and mask, taxonomy and text-role
labels, source and batch metadata, and the other annotator's decisions.

Each annotator selects the referred candidate, assigns the applicable
referential logic, checks unique reference, and judges logic criticality and
shortcut control. Logic labels are mapped deterministically to the five
high-level families used for the primary reproducibility analysis. The audit
records five-family inter-annotator agreement, target-selection accuracy against
the fixed target, and independent pass rates for the three quality criteria.
Target or taxonomy mismatches, inter-annotator disagreements, and negative
quality judgments are resolved by a third adjudicator. Confidence intervals are
computed by resampling the 150 TCESs while keeping the two expressions within
each TCES together.

The completed audit yields \textbf{91.7\%} five-family agreement
(Cohen's $\kappa=\textbf{0.89}$; TCES-clustered 95\% CI:
\textbf{88.3--94.7\%}). At the finer nine-category level, agreement is
\textbf{89.3\%} (Cohen's $\kappa=\textbf{0.88}$; TCES-clustered 95\% CI:
\textbf{85.7--92.7\%}). The annotators also agree on selected target identity
for \textbf{95.7\%} of expressions. Their selections match the fixed target on
\textbf{95.3\%} and \textbf{98.3\%} of expressions, respectively. Their
independent positive rates are \textbf{97.3\%}/\textbf{99.0\%} for unique
reference, \textbf{97.0\%}/\textbf{99.3\%} for logic criticality, and
\textbf{96.0\%}/\textbf{99.7\%} for shortcut control.

\section{GPT Annotation Prompts}
\label{app:gpt_prompt_v22}

The annotation pipeline uses two model calls separated by a
deterministic quota-aware assignment stage. The analysis call receives the
complete image, binary target mask, and known target category and returns
feasible referential logics under a structured JSON schema. The local scheduler
preferentially selects two supported logic categories from different high-level
families when feasible while accounting for category deficits. The generation
call then realizes the selected logics as one minimal expression and one
same-target variant.

The released implementation schema and prompts use the field name
\texttt{operation} for referential-logic labels. They also retain separate
\texttt{spatial\_relation} and \texttt{orientation\_relation} generation tags;
evaluation maps both deterministically to the canonical
\texttt{single\_anchor\_relation} category while preserving the original tags
as auxiliary metadata. The listings below reproduce the prompts verbatim;
angle-bracketed fields denote per-example inputs. The JSON response schemas are
supplied separately through structured output.

\subsection{Stage 1: feasibility analysis}
\begin{Verbatim}
ROLE AND OUTCOME
You are constructing a high-quality instance-level referring-expression segmentation benchmark. Image B fixes
one PRESENT target in Image A. Inspect the full scene and internally test candidate referential operations. Normally
return a compact menu of two or three genuinely feasible operations from different high-level families. Return four
when all four are independently strong and natural, or one only if no reliable cross-family pair exists. Do not
generate final expressions and do not reveal your reasoning.

MANDATORY INTERNAL CHECKS
Before returning an operation, silently verify all of the following:
1. Target preservation: it is visibly true of exactly the masked target and uses no property inferred from the mask.
2. Uniqueness: it rejects every visible same-class competitor, including the hardest competitor you identify
   internally.
3. Minimal logic: removing the defining operation makes the description ambiguous, changes the target, or changes
   the identification problem.
4. Shortcut control: no easier unary target cue in the planned wording makes the assigned operation unnecessary.
5. Anchor validity: every anchor is visible, independently identifiable, and non-circular; multiple anchors are
   returned only when they are jointly necessary.
6. Naturalness: the operation can be expressed as a concise, natural target-centered noun phrase.
7. Difficulty preference: do not return a trivial directional extremum when a reliable anchor-based,
   compositional, target-evidence, negation, or nontrivial same-class operation is available.

INTERNAL COVERAGE SCAN
Silently consider every taxonomy leaf before choosing the compact menu, including orientation_relation,
multi_anchor_relation, nested_grounding, and negation_or_exclusion. This is a recall check, not a quota command:
return one only when it is visibly supported, unique, operation-critical, shortcut-controlled, and natural.

OUTPUT POLICY
- Do not output evidence narratives, distractor descriptions, or chain-of-thought.
- For each operation, output only its labels, a short semantic blueprint, short anchor noun phrases, three audit
  decisions, and confidence.
- Keep logic_blueprint to at most 18 words and each anchor to at most 8 words.
- Set unique, operation_critical, and shortcut_controlled to true only after the corresponding internal check passes.
- Omit any operation that fails a mandatory check. Do not invent an operation to improve dataset balance.
- Treat extremal as a low-priority fallback. Omit it when at least two reliable non-extremal families are available.
- Ordinal means an explicit non-endpoint rank such as second or third from a stated direction. Never classify
  leftmost, rightmost, topmost, bottommost, first, last, or compound forms such as second-lowest as ordinal.
- Set usable_for_benchmark=true only when at least two non-low-confidence operations from different families pass.
- Give a rejection_reason only when usable_for_benchmark=false; keep it to one short sentence.

BOUNDARIES
- Expression role and referential logic are independent dimensions.
- Scene-frame left/right is spatial; leftmost/rightmost within a same-class set is extremal.
- One sufficient anchor is single-anchor; multiple anchors must be jointly necessary for multi-anchor.
- Mentioning another object does not by itself make an expression relational.
- Negation is valid only when exclusion is necessary for uniqueness; it still selects one present target.
- Use affirmative except for negation_or_exclusion, which uses explicit_negation or implicit_exclusion.
- For non-leaf families use the listed subtype; for leaf families use subtype=none.

REFERENTIAL-LOGIC TAXONOMY

1. same_class_selection
   - ordinal: order same-class instances and select an explicit index.
   - extremal: select the instance at a visible positional or measurable extreme.
   - relative_selection: compare or rank same-class instances by a visible property or distance.
   - subgroup_based: define a visually coherent subgroup before selecting one member.

2. single_anchor_relation
   - spatial_relation: resolve one independent anchor and apply one visible spatial/topological relation.
   - orientation_relation: resolve one independent anchor and use facing, looking, or body orientation.

3. multi_anchor_composition
   - multi_anchor_relation: at least two anchors or relations are jointly necessary.
   - nested_grounding: first resolve a relationally described anchor, then locate the target from it.

4. target_centric_visual_evidence (leaf; subtype=none)
   The necessary evidence is directly visible on or through the target: appearance, part, state, pose,
   action, possession, wearing, or interaction. Do not create auxiliary benchmark subtypes.

5. negation_or_exclusion (leaf; subtype=none)
   Eliminate same-class alternatives by a necessary negative or exclusion condition. The target is present;
   this is not an empty-target query or a multi-instance union.


PER-CATEGORY EXAMPLES AND BOUNDARIES (classification guidance, not templates to copy)
Every taxonomy leaf is illustrated below. The examples show the intended logical operation, but an operation may
be returned only when all mentioned evidence is visible and the operation is necessary to identify the masked target.

1. same_class_selection
- ordinal
  Good: "the second of four bottles from the left"; "the third of five birds from the right".
  Boundary: it must use a non-endpoint indexed rank among at least three candidates. "the first bottle from the
  left", "the last bird", and "the second-lowest airplane" are not valid ordinal examples for this benchmark.
- extremal
  Good: "the leftmost bottle"; "the highest bird"; "the largest suitcase".
  Boundary: it selects an endpoint or measurable extreme. It is a low-priority fallback, not a preferred easy
  expression, and it has no independently identified anchor.
- relative_selection
  Good: "the dog larger than the dog beside the door"; "the chair closer to the table than the matching chair by
  the wall".
  Boundary: it compares same-class candidates by a visible property or distance. An explicit second/third index is
  ordinal; a global endpoint such as "the largest dog" is extremal.
- subgroup_based
  Good: "the middle chair among the three chairs around the small table"; "the standing person nearest the two
  seated people".
  Boundary: the subgroup definition must be necessary. "the leftmost chair" alone is extremal, and "the chair by
  the table" alone is a single-anchor spatial relation.

2. single_anchor_relation
- spatial_relation
  Good: "the bottle to the left of the red cup"; "the cat under the wooden table".
  Boundary: the relation is evaluated against one independently identifiable anchor. "the leftmost bottle" has no
  anchor and is extremal; if two anchors are jointly necessary, use multi_anchor_relation.
- orientation_relation
  Good: "the horse facing the red fence"; "the airplane pointing toward the control tower".
  Boundary: facing, gaze, heading, or body orientation relative to one independently identifiable anchor must be
  necessary. "the horse left of the red fence" is spatial, while "the left-facing horse" has no anchor and is
  target-centric visual evidence.

3. multi_anchor_composition
- multi_anchor_relation
  Good: "the person between the woman in red and the man with an umbrella"; "the suitcase below the bench and to
  the right of the blue bin".
  Boundary: both anchors or relations must be jointly necessary. If either anchor alone uniquely identifies the
  target, use the corresponding single-anchor relation instead.
- nested_grounding
  Good: "the dog beside the woman holding an umbrella"; "the chair next to the table behind the sofa".
  Boundary: first resolve an anchor through its own modifier or relation, then use that anchor to locate the target.
  A directly identifiable anchor gives a single-anchor relation; two parallel target-to-anchor constraints give a
  multi_anchor_relation rather than nested grounding.

4. target_centric_visual_evidence (subtype=none)
  Good: "the bird with its wings spread"; "the suitcase with the broken handle"; "the person holding a tennis
  racket".
  Boundary: the decisive appearance, part, state, pose, action, possession, wearing, or interaction evidence is
  visible on or through the target. "the horse facing the red fence" is anchor-relative orientation, and "the
  person beside the bicycle" is a spatial relation rather than target-centric evidence.

5. negation_or_exclusion (subtype=none)
  Good: "the person not wearing a hat"; "the cup without a lid"; "the surfboard that is neither pink nor yellow".
  Boundary: the negative or exclusion condition must eliminate visible same-class alternatives and uniquely select
  one present target. It must not ask for an absent object, return the complement of a region, or denote a union of
  multiple surviving instances.

These examples define boundaries only. Never copy an example unless every mentioned object, property, rank, and
relation is visibly true in the current image and uniquely identifies the masked target.

DYNAMIC INPUT
Known target category: "<KNOWN_TARGET_CATEGORY>". Treat it as a hard constraint. If the image and mask clearly contradict
it, set category_consistent=false and usable_for_benchmark=false. Return only the supplied structured output.
\end{Verbatim}

\subsection{Stage 2: expression generation}
\vspace{-0.5\baselineskip}
\begin{Verbatim}
ROLE AND OUTCOME
Generate exactly two concise noun-phrase referring expressions for the single PRESENT target fixed by Image B:
one minimal expression and one same-target variant. The assigned operations belong to different high-level
families and must identify exactly the same masked instance. Reinspect Image A and perform all checks internally;
do not reveal reasoning or evidence narratives.

MANDATORY INTERNAL CHECKS
- Both expressions preserve the masked target and reject every visible same-class competitor.
- The minimal expression is the shortest natural realization sufficient under primary_logic.
- The variant faithfully realizes variant_logic and is not merely a paraphrase.
- Removing each assigned operation makes its expression ambiguous or changes the identification problem.
- No added color, appearance, or absolute-position cue makes the assigned operation unnecessary.
- Avoid leftmost/rightmost/topmost/bottommost wording unless the assigned subtype is extremal. Even for extremal,
  use it only because the scheduler explicitly selected that low-priority fallback.
- If the assigned subtype is ordinal, use a genuine non-endpoint indexed rank (normally second or later) and never
  use first, last, any superlative, or a compound ranking such as second-lowest.
- Anchors are visible, independently identifiable, and non-circular.
- No expression mentions masks, annotations, boxes, coordinates, panels, or image regions.
- No invisible detail is inferred; negation/exclusion never becomes an absent-target query or union.

OUTPUT POLICY
Return only the two expressions, their assigned labels, and compact boolean self-checks. If either expression fails
any internal check, return needs_reassignment with no expressions and one short rejection reason. Do not weaken,
relabel, or replace an invalid assigned operation.

SELECTED TARGET AND OPERATIONS
{
  "target_head": "<TARGET_HEAD>",
  "primary_logic": "<PRIMARY_LOGIC_JSON>",
  "variant_logic": "<VARIANT_LOGIC_JSON>"
}

RELEVANT CLASSIFICATION BOUNDARIES
- ordinal uses a genuine non-endpoint index among at least three candidates.
- extremal uses an endpoint or superlative and has no independently identified anchor.
- One sufficient anchor is single-anchor; two parallel necessary anchors are multi_anchor_relation; resolving a
  modified or relational anchor first is nested_grounding.
- Negation must uniquely select one present target; it is never an absent-target query or a multi-instance union.
- Never turn a non-extremal assigned operation into an easy extremal phrase.

ASSIGNED-LOGIC REALIZATION EXAMPLES (guidance only; never copy unsupported details)
- ordinal: "the second of four bottles from the left"
- extremal: "the largest suitcase"
- relative_selection: "the dog larger than the dog beside the door"
- subgroup_based: "the middle chair among the three chairs around the small table"
- spatial_relation: "the bottle to the left of the red cup"
- orientation_relation: "the horse facing the red fence"
- multi_anchor_relation: "the person between the woman in red and the man with an umbrella"
- nested_grounding: "the dog beside the woman holding an umbrella"
- target_centric_visual_evidence: "the bird with its wings spread"
- negation_or_exclusion: "the person not wearing a hat"
These examples illustrate operations, not reusable wording. Realize only the assigned logic_blueprint and visible
anchors for the current masked target.

Return only the supplied structured output.
\end{Verbatim}

\section{Training-Regime Audit}

Table~\ref{tab:training_audit} summarizes the downstream supervision associated
with representative evaluated checkpoints. Universal systems combine
RefCOCO-family supervision with mixtures of generic segmentation, reasoning
segmentation, grounded conversation, region question answering, interactive
segmentation, or video data. R+ systems add narrower auxiliary segmentation or
grounding data, while R systems use task-focused referring supervision. CoHD
uses generalized RES supervision, and RefChess composes pretrained components
without task-specific training. The heterogeneous regimes make
Table~\ref{tab:strict_overall} a transfer-performance comparison; controlled
conclusions rely on the matched-supervision study.

\begin{table}[t]
\centering
\caption{Training-regime audit for representative evaluated checkpoints. ``Extra'' indicates downstream supervision beyond classic RefCOCO-family RES.}
\label{tab:training_audit}
\setlength{\tabcolsep}{4pt}
\renewcommand{\arraystretch}{1.08}
\scriptsize
\begin{tabular}{@{}p{1.7cm}p{0.9cm}p{1.5cm}p{1.2cm}p{7.3cm}@{}}
\toprule
Model & Group & Language backbone & Extra & Audited downstream training scope \\
\midrule
X2SAM & U & Qwen3-VL & Yes & RefCOCO/+/g plus generic, reasoning, grounded-conversation, interactive, and visual-grounded segmentation over images and videos. \\
SAMTok & U & MLLM & Yes & Large-scale mask-tokenizer pretraining and mask--text conversations spanning RES, instance/scene parsing, region QA, and interactive reasoning. \\
UniPixel & U & MLLM & Yes & Joint image/video referring, reasoning, video segmentation, captioning, and pixel-QA-related tasks. \\
Sa2VA & U & Qwen3-VL & Yes & Image/video instruction tuning for chat, RES, referring-video segmentation, and grounded caption generation. \\
X-SAM & U & MLLM & Yes & Conversation alignment followed by generic, referring, reasoning, grounded-conversation, interactive, and visual-grounded segmentation. \\
HyperSeg / InstructSeg & U & MLLM & Yes & Joint generic/referring/reasoning and image/video segmentation mixtures. \\
PSALM & U & MLLM & Yes & Caption alignment, panoptic segmentation, RefCOCO/+/g, interactive segmentation, and instruction tuning. \\
ConverSeg & R+ & Qwen2.5-VL & Yes & RefCOCO/+/g plus a synthetic conversational image-segmentation curriculum. \\
EVF-SAM2 / SaFiRe / SSP-SAM & R+ & Mixed / none & Yes & RefCOCO-family training augmented with generic, part, stuff, REC, or grounding datasets, depending on the checkpoint. \\
PaDT / DETRIS / VATEX / LAVT & R & Task-specific & No & RefCOCO-family mixed or dataset-specific referring supervision without unrelated downstream tasks in the evaluated checkpoint. \\
CoHD & G & Specialist & Different & gRefCOCO single-, multi-, count-, and no-target supervision. \\
RefChess & Z & VLM scorer & None & Training-free composition of pretrained SAM and vision--language proposal scoring/search. \\
\bottomrule
\end{tabular}
\end{table}

\section{Additional Experimental Details and Dataset Statistics}
\label{app:extended_results}

This section collects evaluation details and full results omitted from the main
paper for space. Sections~F.1--F.2 document the MLLM and matched-supervision
protocols; Sections~F.3--F.8 provide complete logic profiles, interventions,
dense-track and source analyses, dataset composition, and robustness checks.

\subsection{General-Purpose MLLM Evaluation}

\paragraph{Evaluation protocols.}
The evaluated model families comprise Qwen2.5-VL~\citep{bai2025qwen25vl},
Qwen3-VL~\citep{bai2025qwen3vl}, Qwen3.8-27B~\citep{qwen2026qwen38},
InternVL3.5~\citep{wang2025internvl35},
LLaVA-OneVision~\citep{li2024llavaonevision},
GLM-4.1V~\citep{glmteam2025glm41v},
MiniCPM-V-4.6~\citep{openbmb2026minicpmv46}, and
CogVLM-Grounding~\citep{wang2023cogvlm}, together with the proprietary
Gemini~3.8 Flash~\citep{google2026gemini38} and GPT-5.6
Sol~\citep{openai2026gpt56} APIs in the candidate-ID protocol.
Direct localization presents the original image and one core expression, and
requests exactly one bounding box in a fixed JSON schema. Qwen2.5-VL outputs
absolute image coordinates, while the remaining open checkpoints use integer
coordinates normalized to $[0,1000]$; coordinates are converted to the original
image frame, clipped to its bounds, and reordered if necessary. Open-weight
models use greedy decoding, and closed APIs use structured JSON generation with
temperature zero where supported. An unparseable or invalid box is treated as
an empty prediction and receives zero overlap. Each valid box is passed to the
same frozen \texttt{facebook/sam3} \texttt{sam3.pt} checkpoint as a geometric
prompt. We select SAM~3's highest-confidence proposal, bilinearly upsample its
logits to the original resolution, and threshold the sigmoid mask at $0.5$.
Default output limits range from 128 to 1,024 tokens by checkpoint; only
truncated Qwen3-VL-8B outputs were selectively regenerated with a 4,096-token
limit, leaving already completed predictions unchanged.

The oracle candidate-ID protocol is restricted to $\mathcal{T}_{\mathrm{ID}}$.
Every instance in $\mathcal{C}_t$ is enclosed by an identical yellow box and
assigned a deterministic, target-independent integer ID obtained from a
SHA-256 ordering of the image--category inventory. The model receives the full
scene, the expression, the category, and the valid ID range, and returns one
integer. Local checkpoints use deterministic decoding; API models use the same
prompt with schema-constrained output. Missing, unparsable, or out-of-range IDs
are scored as incorrect. Ground-truth masks remain private scoring data in both
protocols, and the correct candidate ID is additionally withheld under
candidate-ID evaluation.
\begin{table}[t]
\centering
\caption{Direct localization by general-purpose MLLMs on the InstanceBench test
set. Each predicted box is converted to a mask by the same frozen SAM3 backend.
The resulting two-component pipeline is diagnostic and is not directly
comparable with native-mask RES models. All values are percentages.}
\label{tab:mllm_direct}
\setlength{\tabcolsep}{6pt}
\renewcommand{\arraystretch}{1.04}
\small
\begin{tabular}{@{}lrrrrrr@{}}
\toprule
& \multicolumn{3}{c}{Localization and mask quality}
& \multicolumn{3}{c}{TCES reliability} \\
\cmidrule(lr){2-4}\cmidrule(l){5-7}
Model & Box Acc.@.5 & oIoU & mIoU & All@.7 & All-target-ID & Variant ret. \\
\midrule
Qwen2.5-VL-7B & 66.91 & 43.45 & 53.20 & 42.21 & 52.27 & 77.59 \\
Qwen2.5-VL-32B-AWQ & 73.57 & 49.20 & 58.98 & 48.99 & 59.73 & 81.42 \\
Qwen2.5-VL-72B-AWQ & 75.64 & 52.07 & 61.01 & 51.46 & 61.44 & 81.73 \\
Qwen3-VL-4B & 73.83 & 53.90 & 63.20 & 53.99 & 59.94 & 78.85 \\
Qwen3-VL-8B-Thinking & 71.36 & 55.56 & 60.84 & 52.04 & 57.67 & 79.14 \\
Qwen3.8-27B & 81.43 & 60.43 & 69.83 & 63.48 & 69.77 & 85.40 \\
InternVL3.5-8B & 63.04 & 45.96 & 52.08 & 34.94 & 54.81 & 75.83 \\
LLaVA-OneVision-7B & 7.85 & 4.79 & 4.59 & 0.65 & 6.56 & 38.52 \\
GLM-4.1V-9B-Thinking & 71.94 & 51.18 & 61.68 & 52.79 & 57.78 & 78.98 \\
MiniCPM-V-4.6 & 57.21 & 39.70 & 48.08 & 34.81 & 44.94 & 70.51 \\
CogVLM-Grounding & 67.13 & 45.68 & 57.65 & 46.69 & 51.26 & 72.88 \\
\bottomrule
\end{tabular}
\end{table}

\begin{table}[t]
\centering
\caption{Oracle same-class candidate-ID diagnosis on the InstanceBench test set.
The protocol supplies the source-annotated candidate inventory
$\mathcal{C}_t$ and therefore isolates target selection within the annotated
instance space. It is an upper-context
diagnostic rather than the standard RES input protocol. All values are
percentages.}
\label{tab:mllm_candidate_id}
\setlength{\tabcolsep}{6pt}
\renewcommand{\arraystretch}{1.04}
\small
\begin{tabular}{@{}lrrrrr@{}}
\toprule
Model & Expr.-ID & Minimal-ID & Variant-ID & All-target-ID & Variant ret. \\
\midrule
Qwen2.5-VL-7B & 43.95 & 46.69 & 41.21 & 30.61 & 65.57 \\
Qwen2.5-VL-32B-AWQ & 53.38 & 55.09 & 51.67 & 38.91 & 70.63 \\
Qwen2.5-VL-72B-AWQ & 54.11 & 56.03 & 52.20 & 41.63 & 74.30 \\
Qwen3-VL-8B-Thinking & 45.08 & 50.24 & 39.92 & 29.50 & 58.71 \\
Qwen3.8-27B & 66.07 & 68.41 & 63.74 & 54.22 & 79.26 \\
Gemini-3.8-Flash & 76.29 & 77.30 & 75.28 & 65.17 & 84.30 \\
GPT-5.6-sol & 70.76 & 74.09 & 67.43 & 56.52 & 76.28 \\
\bottomrule
\end{tabular}
\end{table}

\subsection{Matched-Supervision Results}

\begin{table}[t]
\centering
\caption{Matched-supervision study on the controlled COCO-derived subset
(mean $\pm$ standard deviation over three seeds). Images, masks, rows,
optimization steps, initialization, and seeds are fixed across supervision
settings; only expression supervision changes.}
\label{tab:matched_training}
\setlength{\tabcolsep}{3.5pt}
\renewcommand{\arraystretch}{1.03}
\small
\begin{tabular}{llrrr}
\toprule
Architecture & Supervision & mIoU & All@.7 & All-target-ID \\
\midrule
\multirow{3}{*}{LAVT}
& Source text & $33.02{\pm}0.69$ & $13.77{\pm}0.54$ & $25.95{\pm}1.00$ \\
& 50/50 mixed & $36.77{\pm}1.49$ & $18.68{\pm}1.91$ & $32.26{\pm}0.78$ \\
& Logic-aware tracks & $38.02{\pm}0.57$ & $20.84{\pm}0.33$ & $33.51{\pm}0.42$ \\
\midrule
\multirow{3}{*}{DETRIS}
& Source text & $43.86{\pm}0.05$ & $29.52{\pm}0.33$ & $37.67{\pm}0.07$ \\
& 50/50 mixed & $45.64{\pm}0.06$ & $32.07{\pm}0.27$ & $42.00{\pm}0.11$ \\
& Logic-aware tracks & $44.79{\pm}0.61$ & $31.27{\pm}1.50$ & $41.19{\pm}1.44$ \\
\bottomrule
\end{tabular}
\end{table}

\paragraph{Matched-supervision implementation.}
Matched supervision uses fixed image-disjoint manifests from a COCO-derived
diagnostic subset. All supervision settings share 930 training images and 1,980
TCESs (7,461 expression--mask rows), together with 84 validation images and 166
TCESs (621 rows). The source-text control replaces every track expression with
its original RefCOCO-family expression; the logic-aware setting retains the
human-verified track expressions; and the 50/50 setting makes a deterministic
row-level mixture of the two. Thus, images, masks, target frequency, row count,
and optimization budget are identical across supervision settings; only
expression supervision changes.

LAVT is initialized from the released RefCOCO checkpoint with BERT-base-uncased
and trained at $480\!\times\!480$ resolution using AdamW, learning rate
$10^{-5}$, weight decay $10^{-2}$, and a polynomial decay schedule with power
$0.9$. DETRIS is initialized from its released mixed checkpoint and official
mixed configuration, and uses Adam with learning rate $2\!\times\!10^{-5}$,
zero weight decay, mixed precision, and the same polynomial schedule. Both
architectures are continued for three epochs with batch size four; the best
validation-mIoU checkpoint is evaluated on the held-out test set. We use
matched seeds 1337, 2027, and 4099 for every architecture and supervision
setting.
\begin{table}[t]
\centering
\caption{Expression-level mIoU by canonical referential-logic category on
the InstanceBench test set. Counts are shared across models.}
\label{tab:logic_profile}
\setlength{\tabcolsep}{2.4pt}
\renewcommand{\arraystretch}{1.03}
\small
\resizebox{\textwidth}{!}{%
\begin{tabular}{lrrrrrrrrr}
\toprule
Model
& \shortstack{Target\\evidence\\(1,273)}
& \shortstack{Single\\anchor\\(1,830)}
& \shortstack{Extremal\\(401)}
& \shortstack{Subgroup\\(321)}
& \shortstack{Multi-\\anchor\\(635)}
& \shortstack{Ordinal\\(403)}
& \shortstack{Exclusion\\(653)}
& \shortstack{Nested\\(207)}
& \shortstack{Relative\\(113)} \\
\midrule
SAMTok & 78.84 & 67.65 & 76.81 & 65.50 & 60.32 & 57.82 & 54.56 & 52.14 & 69.23 \\
X2SAM & 79.19 & 66.00 & 73.75 & 60.59 & 58.66 & 57.92 & 53.95 & 47.84 & 74.07 \\
UniPixel & 75.41 & 64.04 & 73.30 & 57.23 & 57.09 & 45.59 & 51.77 & 46.73 & 64.66 \\
X-SAM & 72.69 & 57.06 & 71.45 & 49.43 & 47.47 & 48.07 & 40.95 & 33.25 & 61.12 \\
Sa2VA & 76.21 & 58.04 & 71.67 & 52.89 & 49.31 & 42.87 & 45.37 & 42.91 & 64.34 \\
ConverSeg & 70.89 & 57.35 & 67.51 & 51.23 & 47.34 & 43.54 & 46.17 & 40.83 & 60.16 \\
EVF-SAM2 & 70.43 & 55.57 & 70.83 & 50.50 & 48.13 & 45.68 & 37.68 & 33.54 & 58.71 \\
DETRIS & 64.36 & 39.45 & 63.30 & 46.23 & 33.28 & 35.93 & 30.13 & 21.64 & 51.56 \\
LAVT-RefCOCO & 49.37 & 27.32 & 58.43 & 41.04 & 24.44 & 31.60 & 19.90 & 16.56 & 40.97 \\
\bottomrule
\end{tabular}}
\end{table}
\begin{figure}[t]
  \centering
  \includegraphics[width=\textwidth]
  {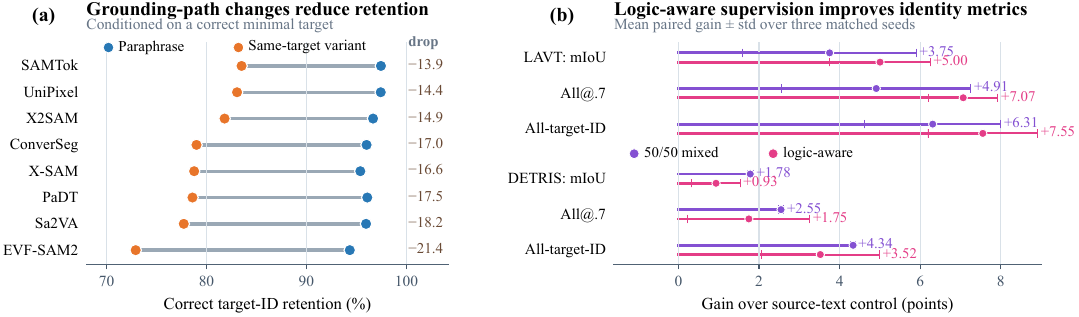}
  \caption{\textbf{Supplementary identity-retention and supervision views.}
  (a) Descriptive target-ID retention conditioned on a correct minimal
  prediction. Each segment connects the same model's paraphrase and
  alternative-path retention; the strictly matched comparison and normalized
  consistency analysis are reported in Table~\ref{tab:identity_consistency}.
  (b) Paired gains over the source-text control in the three-epoch subset study.
  Points are mean within-seed differences and error bars are their standard
  deviations over seeds. Both panels are supplementary:
  panel (a) uses all eligible dense-track instances, while panel (b) measures a
  short continuation rather than general out-of-domain reasoning improvement.}
  \label{fig:supplementary_retention_learnability}
\end{figure}

\subsection{Complete Referential-Logic Profiles}

Table~\ref{tab:logic_profile} gives the full nine-category breakdown behind the
aggregate capability analysis: direct target evidence is consistently easier,
whereas nested, exclusion, and other composition-heavy references remain more
challenging. After matching observable target and expression properties,
Table~\ref{tab:matched_logic_controls} confirms a robust ordinal penalty and a
smaller but generally persistent nested-grounding penalty; the multi-anchor
contrast is weaker and its intervals often include zero.

\begin{table}[t]
\centering
\caption{Matched referential-logic effects in mIoU percentage points. Each
cell reports treated minus control mIoU with a 95\% TCES-clustered bootstrap
confidence interval. Candidate controls are matched exactly by normalized
target category and source stream, then selected by nearest-neighbor distance
over log target-area fraction, log same-class count, and expression length;
controls may be reused. $N$ is the number of matched treated TCESs.}
\label{tab:matched_logic_controls}
\setlength{\tabcolsep}{3.5pt}
\renewcommand{\arraystretch}{1.08}
\scriptsize
\resizebox{\textwidth}{!}{%
\begin{tabular}{lrrrrr}
\toprule
Treated $-$ control & $N$ & X2SAM & SAMTok & UniPixel & X-SAM \\
\midrule
Ordinal $-$ extremal
& 349 & $-10.27\;[-15.82,-4.89]$ & $-17.02\;[-22.65,-11.53]$
& $-25.52\;[-30.51,-20.29]$ & $-20.86\;[-26.23,-15.51]$ \\
Multi-anchor $-$ spatial
& 618 & $-3.13\;[-7.27,1.04]$ & $-1.63\;[-5.66,2.47]$
& $-1.83\;[-5.28,1.78]$ & $-2.17\;[-5.69,1.54]$ \\
Nested $-$ spatial
& 200 & $-13.09\;[-20.43,-6.02]$ & $-7.33\;[-14.45,-0.83]$
& $-6.61\;[-12.90,0.30]$ & $-9.84\;[-15.87,-3.78]$ \\
\bottomrule
\end{tabular}}
\end{table}

\subsection{Controlled Language Interventions}
\label{app:interventions}

\begin{table}[t]
\centering
\caption{Complete controlled-intervention results. Category-only drop is
relative to minimal-expression mIoU. Donor preference measures whether a
same-image, same-category text swap redirects the prediction toward the newly
described instance. The final three columns report paired mIoU changes, in
percentage points, relative to the corresponding original expressions.}
\label{tab:interventions_full}
\setlength{\tabcolsep}{3.2pt}
\scriptsize
\resizebox{\textwidth}{!}{%
\begin{tabular}{lrrrrrrr}
\toprule
& \multicolumn{2}{c}{Category only}
& \multicolumn{2}{c}{Text swap}
& \multicolumn{3}{c}{Inference-time aid ($\Delta$ mIoU)} \\
\cmidrule(lr){2-3}\cmidrule(lr){4-5}\cmidrule(lr){6-8}
Model & mIoU & Drop & Donor IoU & Donor pref.
& Logic-explicit & Deliberation & SC context \\
\midrule
X2SAM & 24.29 & $-44.48$ & 80.61 & 93.50 & $-0.50$ & $+0.13$ & $+0.58$ \\
SAMTok & 24.84 & $-44.91$ & 79.46 & 94.23 & $-0.40$ & $-0.16$ & $-1.52$ \\
UniPixel & 21.24 & $-43.78$ & 77.12 & 93.07 & $-0.29$ & $-0.44$ & $+0.99$ \\
X-SAM & 22.34 & $-36.95$ & 76.60 & 90.29 & $-1.54$ & $-2.05$ & $+2.34$ \\
\bottomrule
\end{tabular}}
\end{table}

Category-only evaluation uses one minimal expression per TCES ($n=2{,}918$),
and the text-swap intervention covers 1,370 expressions with an eligible
same-image, same-category donor. Explicit-logic and candidate-deliberation
prompting cover all 5,836 core expressions; the same-class-context ablation is
restricted to 1,225 identity-eligible same-class-selection expressions. Donor
mIoU reaches 76.60--80.61, confirming that swapped expressions redirect both
target selection and mask output.

We additionally test whether simple inference-time changes repair the diagnosed
weakness. Explicit-logic prompting changes mIoU by $-1.54$ to $-0.29$ points,
candidate deliberation by $-2.05$ to $+0.13$, and oracle same-class context by
$-1.52$ to $+2.34$. These small and model-dependent effects show that the
failure is not reliably corrected by verbalizing the logic, requesting
deliberation, or suppressing non-candidate context.

\subsection{Dense Diagnostic Tracks and Source Transfer}
\begin{table}[t]
\centering
\caption{Track-level mIoU on the 7,370-expression dense test set.
Same-target denotes alternative valid grounding evidence, and exclusion is
grouped by logic category. Paired exclusion drop compares 589
exclusion--minimal pairs from the same TCES. Optional tracks cover different
TCES subsets, so their raw mIoUs are descriptive.}
\label{tab:all_tracks}
\setlength{\tabcolsep}{5.0pt}
\renewcommand{\arraystretch}{1.02}
\small
\begin{tabular}{lrrrrr}
\toprule
Model / checkpoint & Minimal & Paraphrase & Same-target & Exclusion & Paired exclusion drop \\
\midrule
SAMTok-CO & 69.73 & 82.94 & 66.36 & 57.76 & 13.03 \\
X2SAM & 68.74 & 82.22 & 65.08 & 56.90 & 13.81 \\
UniPixel-7B & 65.00 & 80.00 & 62.25 & 54.46 & 13.96 \\
Sa2VA-2B & 62.22 & 78.63 & 56.97 & 48.04 & 17.52 \\
X-SAM & 59.30 & 79.59 & 56.57 & 44.02 & 14.42 \\
ConverSeg-3B & 59.06 & 75.59 & 56.35 & 48.35 & 12.43 \\
EVF-SAM2 & 59.76 & 76.08 & 52.32 & 38.90 & 23.08 \\
PaDT-REC-3B & 43.98 & 70.43 & 43.11 & 31.69 & 9.00 \\
\bottomrule
\end{tabular}
\end{table}
The full eligible dense tracks complement the strictly matched analysis in the
main paper. Table~\ref{tab:all_tracks} shows that alternative-path and exclusion
expressions are less reliable than paraphrases, while
Table~\ref{tab:source_transfer} jointly reports source-specific mIoU and
All@0.7; both descriptive gaps also reflect differences in category and
referential-logic composition. Table~\ref{tab:identity_tracks_full}
reports the corresponding identity-retention results over every eligible
optional track.

\begin{table*}[t]
\centering
\caption{Source-specific expression-level accuracy and TCES-level reliability
on the core test (1,464 COCO and 1,454 OpenImages TCESs; two expressions per
TCES). $\Delta$ is OpenImages minus COCO. X2SAM and X-SAM are three-run means.
Source differences are descriptive because the two subsets also differ in
visual domain, object-category distribution, and referential-logic composition.}
\label{tab:source_transfer}
\setlength{\tabcolsep}{9pt}
\renewcommand{\arraystretch}{1.04}
\small
\begin{tabular}{lrrr@{\hspace{18pt}}rrr}
\toprule
& \multicolumn{3}{c}{Expression-level mIoU}
& \multicolumn{3}{c}{TCES-level All@0.7} \\
\cmidrule(lr){2-4}\cmidrule(lr){5-7}
Model & COCO & OI & $\Delta$ & COCO & OI & $\Delta$ \\
\midrule
SAMTok-CO & 76.86 & 57.35 & $-19.52$ & 75.34 & 43.74 & $-31.60$ \\
X2SAM & 77.80 & 53.96 & $-23.84$ & 75.46 & 37.62 & $-37.84$ \\
UniPixel-7B & 74.61 & 50.86 & $-23.75$ & 70.15 & 33.36 & $-36.79$ \\
Sa2VA-2B & 71.82 & 45.74 & $-26.08$ & 64.69 & 25.93 & $-38.76$ \\
ConverSeg-3B & 70.07 & 43.52 & $-26.56$ & 61.95 & 24.00 & $-37.95$ \\
EVF-SAM2 & 68.45 & 42.20 & $-26.25$ & 56.15 & 21.87 & $-34.28$ \\
X-SAM & 73.64 & 39.86 & $-33.78$ & 67.81 & 25.24 & $-42.56$ \\
InstructSeg & 68.35 & 37.00 & $-31.35$ & 58.33 & 19.67 & $-38.66$ \\
HyperSeg-3B & 68.61 & 36.45 & $-32.16$ & 59.22 & 20.15 & $-39.07$ \\
SaFiRe-mixed & 62.36 & 30.19 & $-32.18$ & 49.32 & 12.93 & $-36.39$ \\
VATEX-Swin-B & 50.20 & 24.39 & $-25.81$ & 32.92 & 10.59 & $-22.33$ \\
\bottomrule
\end{tabular}
\end{table*}

\FloatBarrier
\begin{table}[H]
\centering
\caption{Full dense-track identity retention and target-and-mask retention.
Unlike the strictly matched analysis in Table~\ref{tab:identity_consistency},
each column here uses all TCESs containing that optional track, so these values
are descriptive rather than a controlled path-change comparison.}
\label{tab:identity_tracks_full}
\setlength{\tabcolsep}{5pt}
\renewcommand{\arraystretch}{1.03}
\small
\resizebox{\textwidth}{!}{%
\begin{tabular}{lrrrrrr}
\toprule
Model & Para. Ret. & Variant Ret. & Excl. Ret. & Para. TM@.7 & Variant TM@.7 & Excl. TM@.7 \\
\midrule
SAMTok & 97.45 & 83.51 & 75.57 & 97.12 & 83.64 & 75.17 \\
X2SAM & 96.67 & 81.81 & 71.03 & 96.43 & 81.96 & 72.34 \\
UniPixel & 97.44 & 83.06 & 73.05 & 96.50 & 81.66 & 73.01 \\
Sa2VA & 95.97 & 77.73 & 67.03 & 95.34 & 74.99 & 63.75 \\
ConverSeg & 96.02 & 79.01 & 69.41 & 95.10 & 76.58 & 64.03 \\
PaDT & 96.09 & 78.60 & 68.77 & 95.98 & 82.57 & 71.77 \\
X-SAM & 95.39 & 78.77 & 63.81 & 95.28 & 79.61 & 64.62 \\
EVF-SAM2 & 94.34 & 72.92 & 53.05 & 92.88 & 68.93 & 49.58 \\
\bottomrule
\end{tabular}}
\end{table}

\subsection{Source-Stratified Referential-Logic Robustness}
\label{app:source_logic_robustness}

\begin{figure}[!htbp]
  \centering
  \includegraphics[width=\textwidth]
  {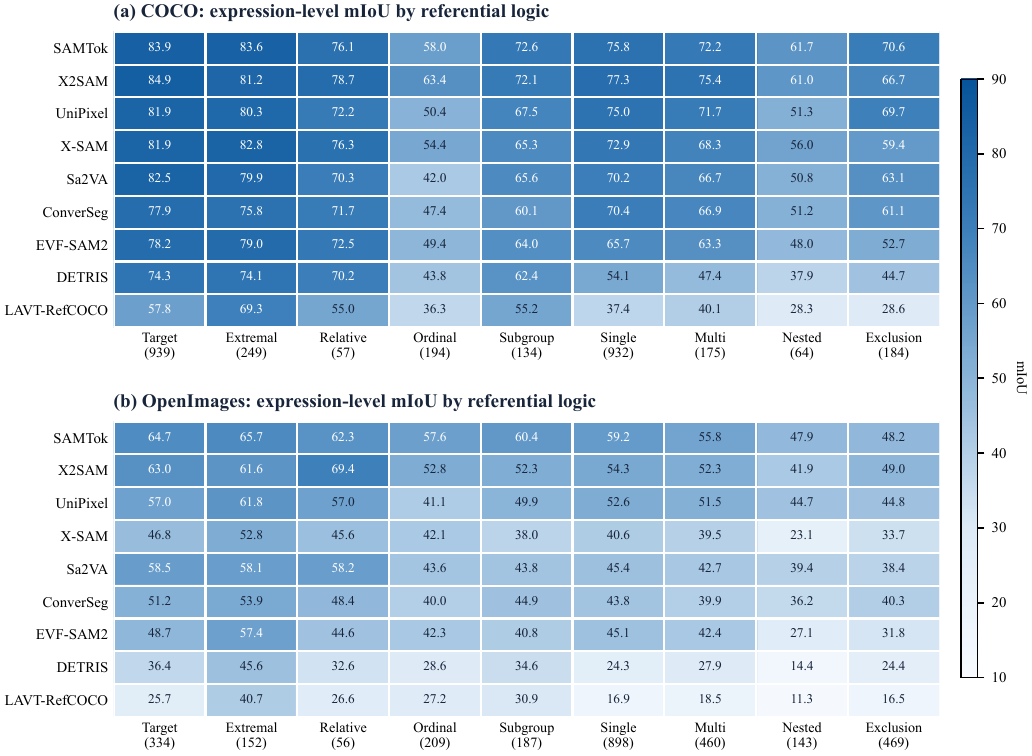}
  \caption{\textbf{Source-specific referential-logic profiles.}
  Cells report expression-level mIoU; parentheses give the number of
  expressions in each source--logic slice.  Across all nine models and in both
  sources, ordinal expressions score below extremal descriptions and nested
  grounding scores below single-anchor relations.  The exact within-source
  ordering is not identical, but the principal difficulty pattern is stable.}
  \label{fig:source_logic_profiles}
\end{figure}

COCO and OpenImages differ in visual domain, object categories, and
referential-logic composition, so a pooled logic profile could in principle
conflate source difficulty with logic difficulty.  Figure~\ref{fig:source_logic_profiles}
therefore recomputes the nine-category profiles within each source.  Absolute
performance is lower on OpenImages, consistent with
Table~\ref{tab:source_transfer}, but the relative difficulty structure is
strongly preserved.

Table~\ref{tab:source_logic_rank} quantifies this stability.  Per-model
Spearman correlations between the COCO and OpenImages category rankings range
from 0.750 to 0.950 (median 0.833).  The ordinal--extremal and
nested--single-anchor gaps are negative for every model in both sources.
Target-centric evidence and extremal descriptions remain among the easier
categories, although which one ranks first can change across models and
sources; we therefore do not claim an identical total ordering.

\begin{table}[!htbp]
\centering
\caption{Cross-source stability of referential-logic difficulty.  $\rho$ is
the Spearman correlation between each model's nine-category mIoU rankings on
COCO and OpenImages.  Gap columns are mIoU percentage-point differences;
negative values mean that ordinal is harder than extremal or that nested is
harder than single-anchor, respectively.}
\label{tab:source_logic_rank}
\setlength{\tabcolsep}{5.0pt}
\renewcommand{\arraystretch}{1.04}
\small
\begin{tabular}{lrrrrr}
\toprule
& & \multicolumn{2}{c}{Ordinal $-$ extremal}
& \multicolumn{2}{c}{Nested $-$ single-anchor} \\
\cmidrule(lr){3-4}\cmidrule(lr){5-6}
Model & $\rho$ & COCO & OI & COCO & OI \\
\midrule
SAMTok & 0.867 & $-25.52$ & $-8.12$ & $-14.10$ & $-11.36$ \\
X2SAM & 0.833 & $-17.78$ & $-8.75$ & $-16.23$ & $-12.37$ \\
UniPixel & 0.950 & $-29.91$ & $-20.69$ & $-23.68$ & $-7.99$ \\
X-SAM & 0.750 & $-28.36$ & $-10.69$ & $-16.87$ & $-17.57$ \\
Sa2VA & 0.833 & $-37.89$ & $-14.49$ & $-19.43$ & $-6.02$ \\
ConverSeg & 0.783 & $-28.45$ & $-13.93$ & $-19.23$ & $-7.60$ \\
EVF-SAM2 & 0.900 & $-29.65$ & $-15.16$ & $-17.66$ & $-18.01$ \\
DETRIS & 0.817 & $-30.33$ & $-16.93$ & $-16.20$ & $-9.90$ \\
LAVT-RefCOCO & 0.767 & $-32.93$ & $-13.48$ & $-9.14$ & $-5.54$ \\
\bottomrule
\end{tabular}
\end{table}

The failure-localization conclusion also holds within each source
(Table~\ref{tab:source_failure_decomp}).  Identity errors account for
66.46--78.79\% of the analyzed COCO failures and 73.50--81.34\% of the
OpenImages failures.  OpenImages nested expressions have a modestly larger
correct-ID/poor-mask share than their COCO counterparts (26.50\% versus
21.21\%), but target-selection errors still dominate that slice.

\begin{table}[!htbp]
\centering
\caption{Source-specific failure decomposition at IoU 0.7, pooled over the
five U-group models SAMTok, X2SAM, UniPixel, Sa2VA, and X-SAM.  $N$ counts
model--expression failures, not unique benchmark expressions.  Same-class
switches are a strict subset of identity errors.}
\label{tab:source_failure_decomp}
\setlength{\tabcolsep}{3.5pt}
\renewcommand{\arraystretch}{1.04}
\small
\begin{tabular}{llrrrr}
\toprule
Source & Logic category & $N$ & Identity error & Same-class switch & Correct ID, poor mask \\
\midrule
COCO & Single-anchor & 984 & 66.46 & 59.45 & 33.54 \\
COCO & Multi-anchor & 205 & 77.56 & 69.76 & 22.44 \\
COCO & Nested & 132 & 78.79 & 70.45 & 21.21 \\
OpenImages & Single-anchor & 2,289 & 76.45 & 51.99 & 23.55 \\
OpenImages & Multi-anchor & 1,211 & 81.34 & 56.48 & 18.66 \\
OpenImages & Nested & 468 & 73.50 & 47.65 & 26.50 \\
\bottomrule
\end{tabular}
\end{table}

\subsection{Supplementary Dataset Composition}

\begin{figure}[!htbp]
  \centering
  \includegraphics[width=\textwidth]
  {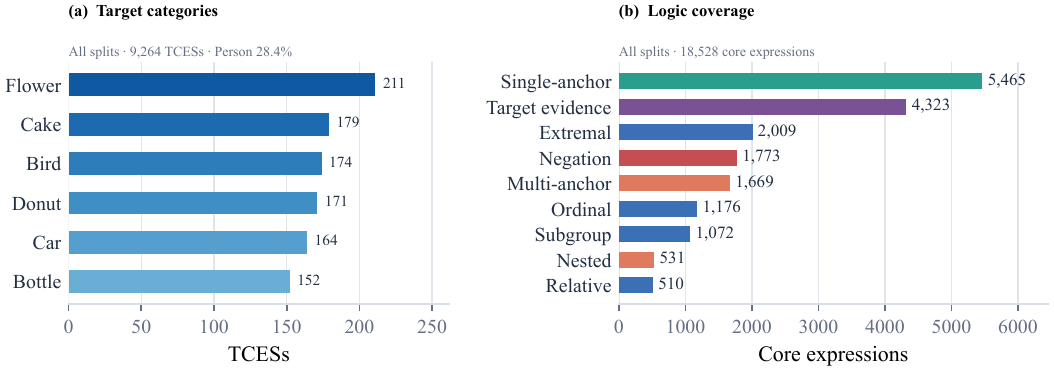}
  \caption{\textbf{Full-dataset composition.}
  (a) The six most frequent non-person target categories, with the person share
  reported in-panel. (b) Coverage of all 18,528 core expressions over the nine
  canonical referential-logic categories. Both panels summarize the complete
  9,264-TCES release.}
  \label{fig:dataset_statistics}
\end{figure}
\begin{figure}[!htbp]
  \centering
  \includegraphics[width=\textwidth]
  {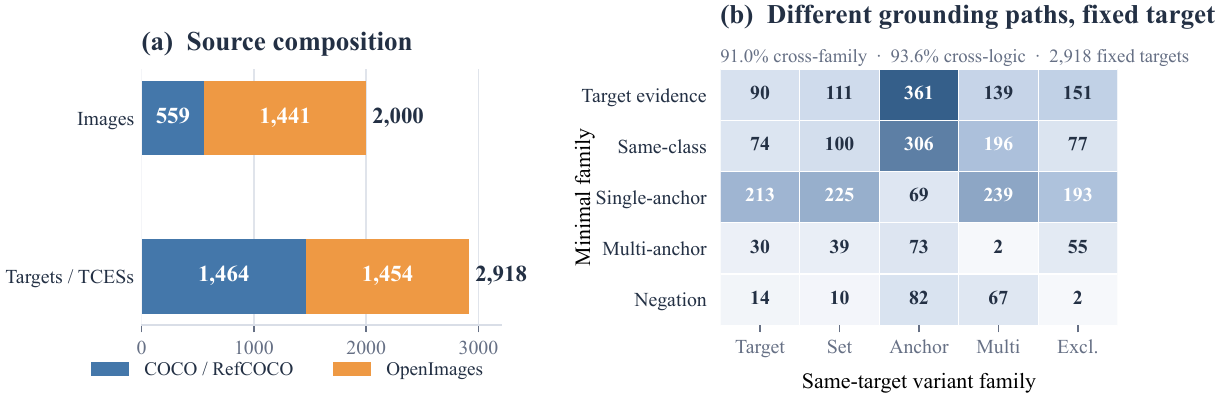}
  \caption{\textbf{Supplementary test-set composition and grounding-path statistics.}
  (a) COCO/RefCOCO and OpenImages contribute complementary image distributions
  and similar numbers of fixed target instances. (b) The full transition
  matrix reports the family of the minimal expression (rows) and same-target
  variant (columns) for all 2,918 test-set TCESs.}
  \label{fig:dataset_statistics_appendix}
\end{figure}

Figure~\ref{fig:dataset_statistics} summarizes target-category and logic
coverage over the complete release, while
Figure~\ref{fig:dataset_statistics_appendix} isolates the test-set source
mix and the grounding-path transition matrix. Together they show that COCO and
OpenImages contribute complementary image distributions and that the benchmark
covers all nine canonical referential logics rather than concentrating on a
single construction pattern.

\subsection{Failure Localization and Statistical Robustness}

\begin{table}[H]
\centering
\caption{Failure decomposition at IoU 0.7, aggregated over the five universal
models in Table~\ref{tab:identity_consistency}. Same-class switches are a strict
subset of identity errors.}
\label{tab:failure_decomp}
\setlength{\tabcolsep}{4pt}
\renewcommand{\arraystretch}{1.03}
\small
\begin{tabular}{lrrr}
\toprule
Logic category & Identity error & Same-class switch & Correct ID, poor mask \\
\midrule
Single-anchor & 73.45 & 54.23 & 26.55 \\
Multi-anchor & 80.79 & 58.40 & 19.21 \\
Nested & 74.67 & 52.67 & 25.33 \\
\bottomrule
\end{tabular}
\end{table}

\begin{table}[!htbp]
\centering
\caption{Sensitivity to the identity-assignment score and margin thresholds,
pooled over the five universal models used in Table~\ref{tab:failure_decomp}.
Core-pair All-target-ID is computed over 14,335 model--TCES pairs. Failure columns use
the 5,289 predictions below IoU 0.7 in the single-anchor, multi-anchor, and
nested categories; identity error is the sum of confirmed same-class switches
and unresolved assignments.  Bold indicates the default setting.}
\label{tab:identity_threshold_sensitivity}
\setlength{\tabcolsep}{5.0pt}
\renewcommand{\arraystretch}{1.04}
\small
\begin{tabular}{rrrrrrr}
\toprule
\shortstack{Min.\\score} & \shortstack{Min.\\margin}
& \shortstack{Core-pair All-\\target-ID}
& \shortstack{Identity\\error}
& \shortstack{Same-class\\switch}
& \raisebox{0.5em}{Unresolved}
& \shortstack{Correct ID,\\poor mask} \\
\midrule
0.05 & 0.00 & 59.77 & 73.28 & 58.04 & 15.24 & 26.72 \\
0.05 & 0.05 & 59.04 & 75.31 & 55.51 & 19.80 & 24.69 \\
0.10 & 0.00 & 59.60 & 73.64 & 57.67 & 15.98 & 26.36 \\
\textbf{0.10} & \textbf{0.05} & \textbf{58.92} & \textbf{75.55}
& \textbf{55.17} & \textbf{20.38} & \textbf{24.45} \\
0.10 & 0.10 & 58.53 & 76.73 & 53.73 & 22.99 & 23.27 \\
0.15 & 0.05 & 58.86 & 75.80 & 54.81 & 20.99 & 24.20 \\
\bottomrule
\end{tabular}
\end{table}

Across all tested settings, identity errors account for 73.28--76.73\% of
relational and compositional failures, while confirmed same-class switches
alone account for 53.73--58.04\%.

\paragraph{Statistical reporting.}
Confidence intervals resample TCESs, preserving within-mask dependence. After
merging the retained spatial and orientation tags, logic-profile counts range
from 113 relative-selection to 1,830 single-anchor expressions; we report every
category but interpret smaller ones through their uncertainty. Raw optional-track
mIoU differences may reflect subset composition, whereas paired identity analyses
use matched TCESs. The NCI analysis retains 988 complete $(m,p,v)$ triplets
after excluding three duplicate-paraphrase TCESs before inspecting model
outputs.

\end{document}